\documentclass[letterpaper, 10 pt, conference]{ieeeconf}  

\usepackage{etoolbox}
\makeatletter
\patchcmd{\@makecaption}
  {\scshape}
  {}
  {}
  {}
\makeatother

\IEEEoverridecommandlockouts                              

\usepackage{mathrsfs}
\usepackage{xcolor}
\usepackage{graphicx}
\usepackage{amsmath}
\usepackage{booktabs}

\usepackage{multirow}
\usepackage{caption}

\usepackage{subcaption}
\usepackage{tikz}
\usepackage{gnuplot-lua-tikz}
\usepackage{svg}
\usepackage{wrapfig}

\usepackage{bm}

\usepackage{amsfonts}

\newcommand{\boldp}[0]{\bm{p}}
\usepackage{csquotes}
\usepackage[breaklinks,colorlinks=true,linkcolor=red, citecolor=blue]{hyperref}%
\usepackage[hyperref=true, backref=false, sorting=none]{biblatex}
\usepackage{cleveref}

\makeatletter
\def\footnoterule{\kern-3\p@
  \hrule \@width 2in \kern 2.6\p@} 
\makeatother

\title{\LARGE \bf
3QFP: Efficient neural implicit surface reconstruction using Tri-Quadtrees and Fourier feature Positional encoding
}

\author{Shuo Sun$^{1}$, Malcolm Mielle$^{2}$, Achim J. Lilienthal$^{1,3}$, and Martin Magnusson$^{1}$
\thanks{*This work has received funding from the European Union’s Horizon 2020 research and innovation programme under grant agreement No 101017274 (DARKO).}
\thanks{$^{1}$AASS MRO lab, \"{O}rebro University, Sweden.
    {\tt\small 
        \{shuo.sun, achim.lilienthal, martin.magnusson\}@oru.se
    }
}%
\thanks{$^{2}$Independent researcher.
        {\tt\small {malcolm.mielle@protonmail.com}}
}%
\thanks{$^{3}$Technical University of Munich, Chair: Perception for Intelligent Systems.
    {\tt\small 
        achim.j.Lilienthal@tum.de
    }
}%
}

\begin{document}

\maketitle
\thispagestyle{empty}
\pagestyle{empty}

\begin{abstract}
  Neural implicit surface representations are currently receiving a lot of interest as a means to achieve high-fidelity surface reconstruction at a low memory cost, compared to traditional explicit representations.
  However, state-of-the-art methods still struggle with excessive memory usage and non-smooth surfaces. This is particularly problematic in large-scale applications with sparse inputs, as is common in robotics use cases.
  To address these issues, we first introduce a sparse structure, \emph{tri-quadtrees}, which represents the environment using learnable features stored in three planar quadtree projections.  
  Secondly, we concatenate the learnable features with a Fourier feature positional encoding.
  The combined features are then decoded into signed distance values through a small multi-layer perceptron.
  We demonstrate that this approach facilitates smoother reconstruction with a higher completion ratio with fewer holes.
  Compared to two recent baselines, one implicit and one explicit, our approach requires only 10\%--50\% as much memory, while achieving competitive quality.
  %
The code is released on \color{blue}{\textbf{$\texttt{https://github.com/ljjTYJR/3QFP}$}}.
\end{abstract}

\section{INTRODUCTION}

Most autonomous systems rely on an accurate model of the environment, i.e. a map, for localization and planning.
Various methods have been designed to represent maps as point clouds~\cite{dellenbach2022cticp,vizzo2023kiss}, spatial voxels with normal distributions~\cite{magnusson-2007-jfr, magnusson2009appearance,stoyanov2010ndtpath} or SDF values~\cite{oleynikova2017voxblox,pan2022voxfield,Canelhas2013SDFTA, CanelhasRAL},
occupancy grids~\cite{hornung2013octomap,duberg2020ufomap},
surfels~\cite{behley2018suma}, meshes~\cite{vizzo2021puma}, etc.
However, these methods often require substantial memory resources to maintain an accurate and detailed environment representation, especially in large-scale scenes. Limiting the available memory, on the other hand, will decrease the map quality.
In addition, these methods encounter difficulties in accurately reconstructing the environment in detail when the input data is sparse, resulting in holes and non-smooth surfaces in the map (see \href{https://www.roboticsproceedings.org/rss14/p16.pdf}{Fig.~4} in \cite{behley2018suma} and \href{https://www.ipb.uni-bonn.de/wp-content/papercite-data/pdf/vizzo2021icra.pdf}{Fig.~1} in \cite{vizzo2021puma}).

Recent neural implicit representation techniques have achieved notable success in shape representation~\cite{park2019deepsdf,mescheder2019occupancy,takikawa2021nglod} and scene representation~\cite{peng2020convolutional}.
These methods enable the implicit storage of environmental information within a neural network and/or within learnable feature volumes, thereby enabling compact yet detailed reconstruction of such environments.
However, previous neural implicit reconstruction techniques primarily focus on objects or small scene reconstruction~\cite{jiang2023h2}, and the memory footprint does not scale to large-scale environments.
 Recent research has explored the use of sparse structures such as octrees~\cite{yu2023nf-atlas, zhong2023shine} for better scalability.
Furthermore, when inputs are sparse, these methods can often lead to the generation of non-smooth surfaces~\cite{johari2023eslam}, a phenomenon that remains a significant challenge in the field.

 \begin{figure}[t!]
     \centering
     \begin{minipage}[t]{0.8\linewidth}
         \includegraphics[width=\linewidth]{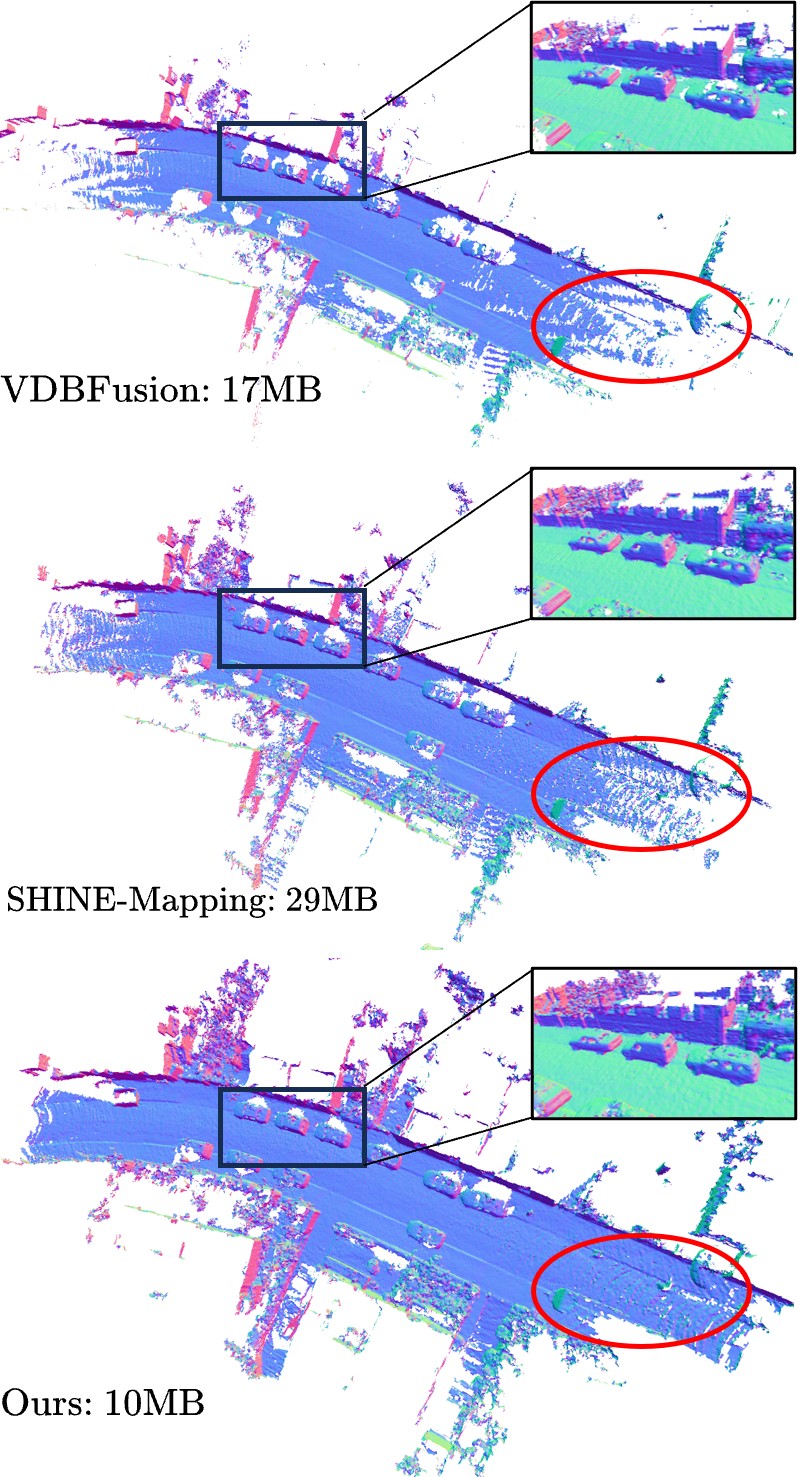}
     \end{minipage}
     \caption{Qualitative reconstruction result on $\texttt{KITTI-Seq07}$.
     Our method can achieve better reconstruction quality using less memory compared to \emph{SHINE-Mapping}~\cite{zhong2023shine} and \emph{VDBFusion}~\cite{vizzo2022vdbfusion}.
     When given noisy and sparse lidar scans, our method can achieve a more complete reconstruction (see {\color{red}{red circle}} and \textbf{black zoomed-in} square areas).
     }
     \label{fig:kitti}
     \vspace{-0.8cm}
\end{figure}

Our first contribution is a novel feature representation structure called \emph{tri-quadtrees}.
Instead of storing features in 3D voxel grids~\cite{zhong2023shine,voxFusion,voxSurf} or dense feature planes~\cite{johari2023eslam}, we use three planar quadtrees to represent surfaces.
Our method combines the sparsity of the octree and the efficiency of the feature planes, greatly reducing memory requirements while still producing comparable results, as shown in \cref{fig:kitti}. 

Our second contribution is a hybrid feature representation method.
Although learnable features can achieve detailed and high-fidelity reconstruction, they tend to degenerate when inputs are sparse~\cite{Yang2023FreeNeRFIF,wang2023coslam}.
We combine the learnable features with the Fourier feature positional encoding~\cite{tancik2020fourier}, which can help fill holes and smoothen reconstruction when given sparse inputs.

The concatenated features are fed into a small multi-layer perceptron~(MLP) for predicting a continuous signed distance field of the scene.
Features and MLP parameters are optimized end-to-end under the supervision of the direct range measurement.
As our experiments demonstrate, compared to the recent explicit SDF/TSDF representation VDBFusion~\cite{vizzo2022vdbfusion}, our method achieves more complete reconstructions; compared to the state-of-the-art neural implicit reconstruction~\cite{zhong2023shine}, our method requires significantly less memory.

\section{Related Work}
\subsection{Explicit Representation}
The most common representations used in current robot mapping implementations are explicit representations, such as point clouds~\cite{dellenbach2022cticp,vizzo2023kiss}, NDTs~\cite{stoyanov2010ndtpath,magnusson2009appearance}, occupancy grids~\cite{duberg2020ufomap,hornung2013octomap}, surfels~\cite{behley2018suma}, meshes~\cite{vizzo2021puma,rosinol2019kierma}, TSDF values~\cite{oleynikova2017voxblox,vizzo2022vdbfusion}, etc.
\emph{VDBFusion}~\cite{vizzo2022vdbfusion} stores TSDF values in sparse voxels explicitly; each voxel will be queried during reconstruction.
Unlike previous approaches that explicitly divide TSDF or SDF maps into voxels, our method adopts a neural network to store SDF values and allows queries of arbitrary coordinates in a continuous manner.
The use of neural networks allows our method to achieve high reconstruction completion ratios while maintaining the smoothness of the reconstructed surface.

\subsection{Neural Implicit Representation}
\subsubsection{Pretrained Encoder-Decoder}
\emph{DeepSDF}~\cite{park2019deepsdf} and \emph{OccNet}~\cite{mescheder2019occupancy} propose to use a neural network to represent the SDF and occupancy probability for the shape.
\emph{CovONet}~\cite{peng2020convolutional} maps 3D points to feature grids based on the PointNet encoder~\cite{Qi_2017_CVPRpointNet}, and then uses an MLP as a decoder for predicting occupancy probability. 
The encoder and decoder are trained in an end-to-end manner.
When querying a coordinate $\boldp$, the feature value at $\boldp$ is interpolated by the spatially adjacent grids and fed to the decoder.
The trained encoders and decoders can be used directly in other environments for inference, which, however, often suffer from the generalization problem~\cite{johari2023eslam}.

\subsubsection{Test-time optimization}
The seminal work of \emph{NeRF}~\cite{mildenhall2021nerf} on novel view synthesis has inspired a surge of research activity in geometry reconstruction.
For example, \emph{iMAP}~\cite{sucar2021imap} and \emph{iSDF}~\cite{ortiz2022isdf} use an overfit large MLP to store environment information, with the environment representation stored within the network parameters.
The training of such large MLPs is, however, time-consuming, prompting the development of various techniques aimed at addressing this issue.
One efficient manner is to move the computational complexity from the neural network to the scene feature volume,
introducing dense feature voxel grids~\cite{Yu2022MonoSDF,gosurf}.
With learnable features, only a small MLP can be used as a decoder, reducing computational load and accelerating training.
Accounting for the large memory footprint when applying dense feature voxel grids,  several techniques have been proposed to reduce memory usage, such as hash-tables~\cite{muller2022instant}, octree-trees~\cite{takikawa2021nglod}; these compact data structures have been leveraged in recent robotic applications~\cite{zhong2023shine,jiang2023h2,yu2023nf-atlas,wang2023coslam,voxSurf,voxFusion,yan2023EINR}.
However, these voxel-based feature representations can still consume $\mathcal{O}(n^{3})$ memory in the worst case.
To further reduce memory usage, recent works~\cite{chan2022eg3d,johari2023eslam,hu2023triMipNerf,reiser2023merf,wang2023petneus,fridovich2023kplane} has proposed using three orthogonal axis-aligned feature planes instead of 3d voxel grids.
  
Motivated by the feature plane representation, we propose a more compact representation called tri-quadtrees, which can further reduce memory usage.

The work similar to ours is \emph{SHINE-Mapping}~\cite{zhong2023shine}.
However, we propose a novel feature representation method that significantly reduces memory usage in large-scale scenes.
In addition, inspired by \emph{CO-SLAM}~\cite{wang2023coslam}, 
we combine Fourier feature positional encoding with learnable features, resulting in smoother and more complete reconstructions when input data is sparse.

\section{Method} \label{sec.methods}

\begin{figure*}
    \centering
    \includegraphics[width=0.8\linewidth]{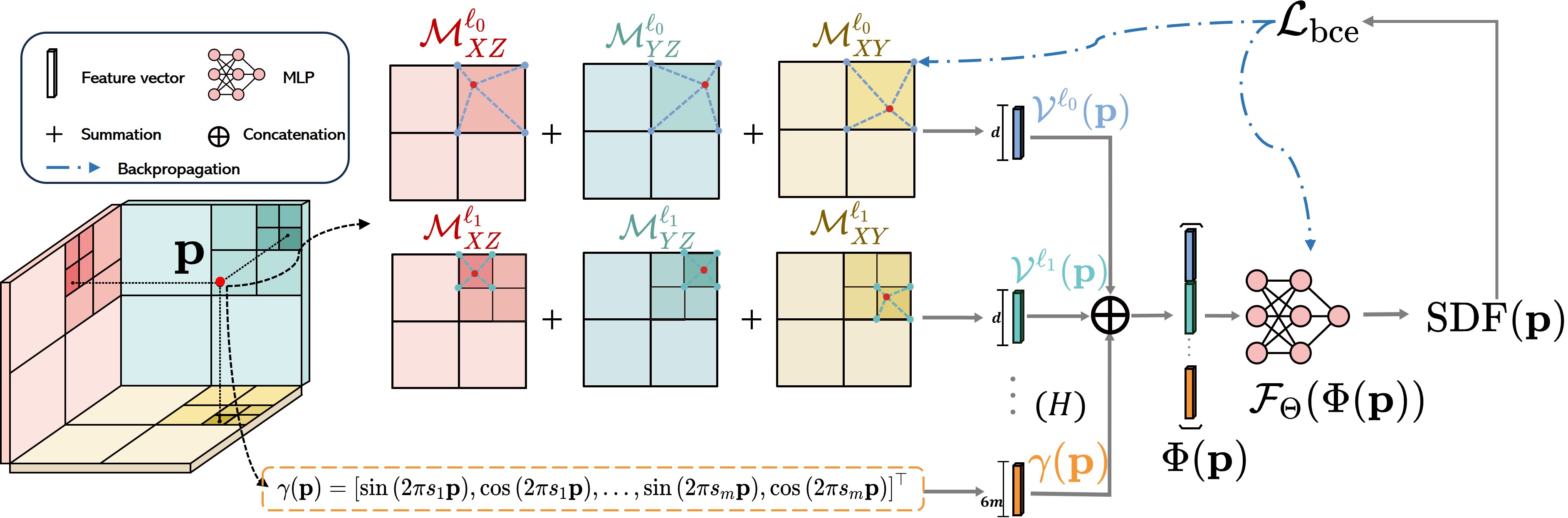}
    \caption{ \textbf{Overview of our method.}
    We represent the scene with three planar quadtrees $\mathcal{M}_{i}^{\ell}$, $i \in \{XZ,YZ,XY\}$ and $\ell$ represents the quadtree depth.
    We store features in the deepest $H$ levels of resolution of quadtrees.
    When querying for a point $\boldp$, we project it onto planar quadtrees to identify the node containing $\boldp$ at the level $\ell$.
    The feature of $\boldp$ is then calculated by bilinear interpolation based on the queried location and vertex features.
    We add features at the same level and concatenate among different levels.
    Concatenated with the positional encoding $\gamma(\boldp)$, $\boldp$'s feature~($\Phi(\boldp)$) is fed into a small MLP~($\mathcal{F}_\Theta$) to predict the SDF value.
    The learnable features stored in the quadtree nodes and the network parameters are learned by test-time optimization using the loss function $\mathcal{L}_{\text{bce}}$.
    The learnable feature vectors have length $d$ and the positional encoding feature vector has length $6m$.
  }
    \label{fig:overview}
    \vspace{-0.5cm}
\end{figure*}

Our method learns a continuous signed distance function representation of the environment given lidar scans and known poses.
Specifically, the world coordinate $\boldp_{i} \in \mathbb{R}^{3}$ is mapped into a SDF value $s_{i} \in \mathbb{R}$.
As shown in \cref{fig:overview}, our neural implicit representation is composed of two components: the learnable features stored in the quadtree nodes and a globally shared MLP for predicting the SDF value.
The features and the network parameters are learned during  test time by using direct lidar measurement to supervise network predictions.

\Cref{sec.tri-quadtrees} delves into how we use the novel \emph{tri-quadtrees} structure to compactly represent a given scene.
\Cref{sec.ffmapping} elaborates on the application of Fourier features positional encoding.
Finally, \cref{sec:training_optimization} introduces the loss function and some training details.
Based on the proposed neural implicit representation,
we extract the mesh by Marching Cubes \cite{marchingcubes} for visualization and evaluation.

\subsection{Tri-Quadtrees feature representation} \label{sec.tri-quadtrees}
\textbf{Tri-Quadtrees:}
While storing learnable features in spatial voxel grids~\cite{gosurf} can lead to fast convergence rates,
dense 3D grids suffer from a cubical growth rate of memory usage as the environment size increases.
To avoid storing unnecessary features in free space, prior work~\cite{zhong2023shine,voxFusion,takikawa2021nglod,voxSurf} employs octree to store features only within voxel grids where surface points are located.
Other research such as \textcite{johari2023eslam} adopts feature planes, which represent scene features by three orthogonal planes.
In this paper, we propose a novel data structure to represent spatial features---tri-quadtrees, which is more compact than previous methods.

Specifically, given lidar scans and known poses, we first project the 3D point clouds onto three axis-aligned orthogonal 2D planes and then construct quadtrees on each plane with a maximum depth of $L_{\text{max}}$ and resolution of the leaf nodes equal to $r$.
Unlike the method in \emph{NGLoD}~\cite{takikawa2021nglod}, which uses all levels of resolution in the octree to store features, 
we only use the deepest $H$ levels of quadtrees for feature representation to balance the quality of the reconstruction and memory footprint.
That is, only the levels $\ell \in \{L_{\text{max}}-H+1, \cdots, L_{\text{max}}\}$ in the quadtree are used to interpolate features.
By default, we set $H=3$, which was empirically sufficient to achieve good results.


\textbf{Feature Computation:} We denote the ``feature plane quadtree" as $\mathcal{M}_{i}^{\ell}$ where $i \in \{XY, XZ, YZ\}$ is the plane index and $\ell \in \{L_{\text{max}}-H+1, \cdots, L_{\text{max}}\}$ is the depth level.
Each node in the depth $\ell$ holds a one-dimensional learnable feature vector $\mathcal{V}_{i}^{\ell,(j)}$ at each of its four vertices (indexed by $j$), where $\mathcal{V}_{i}^{\ell,(j)} \in \mathbb{R}^{d}$ and $d \in \mathbb{N}$ is the length of the feature vector.
The features are initialized randomly when created and optimized until convergence during training.

For quick queries, we store vertex features with hash tables.
Specifically, for each level $\ell \in \{L_{\text{max}}-H+1, \cdots, L_{\text{max}}\}$ of each quadtree, we maintain a hash table (totally $3H$) to store vertex features.
The vertex Morton code is used as hash table keys, which can map two-dimensional vertex indices to one-dimensional scalars.

To get the feature for a query point $\boldp$ at the level $\ell$, we first project $\boldp$ to planes and get the 2D points $\boldp_{x}$, $\boldp_{y}$ and $\boldp_{z}$.
Then we traverse the quadtree at level $\ell$ to find the node containing the corresponding 2D point.
The feature for the 2D point is bilinearly interpolated by the four vertex features in the node. The feature $\mathcal{V}^{\ell}(\boldp)$ of  level $\ell$ for the point $\boldp$ is calculated through summation for all 3 planes: $\mathcal{V}^{\ell}(\boldp) = \mathcal{V}^{\ell}_{XZ}(\boldp_{y}) + \mathcal{V}^{\ell}_{YZ}(\boldp_{x}) + \mathcal{V}^{\ell}_{XY}(\boldp_{z})$, $\mathcal{V}^{\ell}(\boldp) \in \mathbb{R}^{d}$.

The final feature $\mathcal{V}(\boldp)$ will be collected and concatenated across all $H$ levels:
$$
    \mathcal{V}(\boldp) = [\mathcal{V}^{L_{\text{max}} - H + 1}(\boldp), \mathcal{V}^{L_{\text{max}} - H + 2}(\boldp), \cdots, \mathcal{V}^{L_{\text{max}}}(\boldp)]^{\top},
$$
where $[\cdot,\cdot]$ denotes concatenation, $\mathcal{V}(\boldp) \in \mathbb{R}^{dH}$.




\subsection{Fourier features positional encoding} \label{sec.ffmapping}
Although the learnable features can contribute to a more accurate reconstruction,
they do not provide the necessary hole filling and smoothness, as demonstrated by \textcite{wang2023coslam}.
To improve the completion ratio of the reconstruction, we combine the learnable features and the positional encoding.
We demonstrate through experiments~(\cref{sec.abla}) that combining learnable features with the positional encoding can boost the completion ratio with only a marginal increase in computational cost.

In contrast to the setup in~\textcite{wang2023coslam}, we do not employ the \emph{one-blob} encoding.
Instead, we found that Fourier feature positional encoding can achieve a smoother result in our cases.
Inspired by~\textcite{tancik2020fourier}, we adopt Gaussian positional encoding which has been shown to achieve a higher completion ratio than \emph{one-blob} and frequency encoding used in \emph{NeRF}~\cite{mildenhall2021nerf}.
Specifically, the positional encoding for the query point $\boldp$ would be
\begin{align*}
    \gamma(\boldp) = [&\sin{(2\pi s_{1} \boldp)}, \cos{(2\pi s_{1} \boldp)},\\ &\ldots ,\sin{(2\pi s_{m} \boldp), \cos{(2\pi s_{m} \boldp)}}]^{\top}
\end{align*}
where $s_{i}$ are coefficients  ($i \in \{1,2,\ldots,m\}$) sampled from an isotropic Gaussian distribution, i.e., $s_{i}\sim \mathcal{N}(0,\sigma^{2})$, where $\sigma^{2}$ is a tuned hyperparameter; $m$ is a hyperparameter controlling the length of the positional encoding feature.
We determine $\sigma^{2}$ by doing a hyperparameter search based on the training loss.

Finally, for the query point $\boldp$, we concatenate the tri-quadtrees feature $\mathcal{V}(\boldp) \in \mathbb{R}^{dH}$ and Fourier positional encoding $\gamma(\boldp) \in \mathbb{R}^{6m}$ into the final feature vector $\Phi(\boldp)=[\mathcal{V}(\boldp), \gamma(\boldp)]^{\top} \in \mathbb{R}^{dH+6m}$.

To query the SDF value, we feed the feature $\Phi(\boldp)$ into a small MLP $\mathcal{F}_{\Theta}$ to obtain the SDF value, where $\Theta$ represents the network parameters.
The whole process is differentiable, we jointly optimize the quadtree features and the MLP parameter $\Theta$ end-to-end during training.

\subsection{Training and Optimization} \label{sec:training_optimization}
\subsubsection{Sampling}
We take samples from both free-space and points close to the surface.
For each lidar ray, we randomly select $N_{s}$ points $\{\boldp_{i}\}_{i=1}^{N_{s}}$ in a truncated area of the ray's endpoint $\bm{p}_{e}$; we also take $N_{f}$ points $\{\boldp_{j}\}_{j=1}^{N_{f}}$ in free space along the ray.
For SDF supervision signals, we directly calculate the distance between the sampled points $\bm{p}_{s}$ and the endpoint $\bm{p}_{e}$\footnote{Other supervision signals such as the calculated gradients~\cite{wiesmann2023locNDF} and the nearest distance~\cite{ortiz2022isdf} can also be used. Since lidar often provides accurate range measurement compared to RGB-D cameras, we adopt such \emph{projected distance} directly~\cite{oleynikova2017voxblox,zhong2023shine}}.
The SDFs of the points located between the sensor and the endpoint will have a negative value, while those outside will be positive.

\subsubsection{Loss Function}
Our training only needs a simple SDF loss function\footnote{We also attempted to include additional regularization losses such as the Eikonal loss~\cite{Yu2022MonoSDF}, however, it did not make a significant difference; therefore, we decided to omit it for the sake of simplicity.}.
Following the configuration in \cite{zhong2023shine,Yang2023FreeNeRFIF}, we adopt the Binary Cross Entropy~(BCE) loss for faster convergence.
Specifically, we map both the ground truth $\text{SDF}_{\text{gt}}$ and the predicted $\text{SDF}_{\text{pred}}$ to the range $[0,1]$ using the sigmoid function (\texttt{Sig}).
For one training pair, the BCE loss will be:
\begin{align}
    \mathcal{L}_{\text{bce}} = o_{\text{gt}} \cdot \log{(o_{\text{pred}})} + (1 - o_{\text{gt}}) \cdot \log{(1-o_{\text{pred}})},
\end{align}
where $o_{\text{gt}}=\texttt{Sig}(\text{SDF}_{\text{gt}})$ and $o_{\text{pred}} = \texttt{Sig}(\text{SDF}_{\text{pred}})$ are the ground truth and predictions respecitvely.

\section{Experiments}
In this section, we first demonstrate through quantitative experiments that tri-quadtrees is more memory efficient than the current state-of-the-art implicit representation methods (see \cref{sec.mem_eff}) while still achieving a higher completion ratio than explicit representation methods, as shown in \cref{sec.map_qual}.
Qualitative results demonstrate that our method is capable of producing smoother reconstructions when inputs are sparse.
Finally, an ablation study reveals that the use of positional encoding can significantly enhance smoothness and hole-filling~(\cref{sec.abla}).

\subsection{Experiment Setup}
\textbf{Baseline:}
We compare our approach to the state-of-the-art explicit reconstruction method \emph{VDBFusion}~\cite{vizzo2022vdbfusion} and the neural implicit reconstruction method $\emph{SHINE-Mapping}$~\cite{zhong2023shine}, both of which have publicly available implementations and are based on TSDF or SDF representations.

\textbf{Evaluation Metric:}
Following the setup of the experiment in ~\cite{vizzo2021puma, zhong2023shine}, we evaluate the geometry of the reconstruction by indirectly examining the mesh constructed from the TSDF output (created by \emph{marching cubes}~\cite{marchingcubes}).
We uniformly sample $10^{7}$ points using $\texttt{open3d}$ library from the generated meshes and compare with the ground truth point cloud for evaluation.
We report accuracy [cm], completion [cm], completion ratio [$\%$], and accuracy ratio [$\%$]. Briefly, let $\mathcal{P}$ be the point cloud sampled from the prediction mesh, while $\mathcal{G}$ is the ground truth point cloud. For a point $\bm{p}_{i} \in \mathcal{P}$, we define the distance to $\mathcal{G}$ as
$$
d(\bm{p}_{i}, \mathcal{G}) = \min_{\bm{g}_{i} \in \mathcal{G}} ||\bm{p}_{i} - \bm{g}_{i}||.
$$
Similarly, the distance between the point $\bm{g}_{i}$ and $\mathcal{P}$ is
$
d(\bm{g}_{i}, \mathcal{P}) = \min_{\bm{p}_{i} \in \mathcal{P}} ||\bm{g}_{i} - \bm{p}_{i}||.
$
We compute $d(\bm{p}_{i}, \mathcal{G})$ as accuracy and $d(\bm{g}_{i}, \mathcal{P})$ as completion, the ratio of which less than set threshold works as accuracy ratio and completion ratio.
In addition, we measure the memory usage of each method.

\textbf{Datasets}:
We evaluate our approach on two public lidar datasets.
One is $\texttt{MaiCity}$\footnote{\href{https://www.ipb.uni-bonn.de/data/mai-city-dataset/}{https://www.ipb.uni-bonn.de/data/mai-city-dataset/}}---a synthetic urban-like outdoor scenario with
100 lidar frames.
The other one is $\texttt{NewerCollege}$~\cite{ramezani2020newercollege}, a real lidar dataset including 1500 frames captured in a college campus environment.
The two datasets provide registered dense point clouds as ground truth reference for quantitative evaluation.
\Cref{fig:kitti} includes a qualitative comparison on the KITTI dataset, but since KITTI does not provide accurate ground truth we do not compare quantitative accuracy numbers on that dataset.

\textbf{Implementation Details:}
In all experiments, our method uses the parameters shown in \cref{tab.our_parameters} (we use these parameters to trade off efficiency and accuracy). We set $\sigma^{2}=50$ by parameter searching on \texttt{MaiCity} and apply it on both datasets.

\begin{table}
\centering
\captionsetup{position=top}
\caption{Parameter Setting: the table shows the important hyperparameters appearing in \cref{sec.methods}.}
\begin{tabular}{ccc|ccc}
\toprule
\textbf{Category}   &   \textbf{Parameter}  &   \textbf{Value}  &   \textbf{Category}   &   \textbf{Parameter}  &   \textbf{Value} \\
\cmidrule(r){1-3} \cmidrule(l){4-6}
\multirow{2}{*}{Quadtree}   &   $H$ &   3   &   \multirow{2}{*}{Feature}    &   $d$ &   8   \\
&  $L_{\text{max}}$ & 12 & & $m$ & 16 \\
\cmidrule(r){1-3} \cmidrule(l){4-6}
\multirow{2}{*}{MLP} & depth & 2 & \multirow{2}{*}{Sampling} & $N_{s}$ & 3\\
& hidden\_width & 32 & &$N_{f}$&3\\
\bottomrule
\label{tab.our_parameters}
\end{tabular}
\vspace{-0.5cm}
\end{table}

We employ the default or recommended configurations for the baseline methods.
Specifically, we use 0.1~m leaf node resolution in the $\emph{SHINE-Mapping}$ octree and our tri-quadtrees.
Correspondingly, we use 0.1~m voxel size for $\emph{VDBFusion}$.
In the experiment, we use marching cubes with  0.1~m voxel resolution to extract meshes for all methods.

\subsection{Map Quality} \label{sec.map_qual}

In our first map quality evaluation experiment, we use a \textit{dense} input configuration:
in \texttt{MaiCity} all frames are used, while \texttt{NewerCollege} selects one frame out of every three, to show that our method can achieve competitive performance with a smaller memory footprint.

The evaluation results are shown in \cref{tab.dense_mapping}.
While \emph{VDBFusion} achieves slightly better accuracy than the implicit methods, the completion metric is much lower than the neural implicit reconstruction methods.
Our method and \emph{SHINE-Mapping} have a significantly higher completion ratio with only a slight drop in accuracy.

\begin{table*}
\centering
\caption{Quantitative evaluation of the reconstruction quality on the $\texttt{MaiCity}$ and $\texttt{NewerCollege}$ with $\emph{dense}$ inputs.
We report the \emph{Completion}~(Comp.), \emph{Accuracy}~(Acc.), \emph{Completion Ratio}~(Comp.Ratio) and \emph{Accuracy Ratio}~(Acc.Ratio) with a threshold of 0.1~m for $\texttt{MaiCity}$ and 0.2~m for $\texttt{NewerCollege}$.
We also report the number of learnable parameters for neural implicit representation methods.
Bold fonts represent the best results.
Our method achieves a significantly higher completion ratio than \emph{VDBFusion} with fewer parameters than \emph{SHINE-Mapping}.
($\downarrow$: lower better; $\uparrow$: higher better.)
}
\begin{tabular}[t]{*{7}{c}}
\toprule
\textbf{Dataset} & \textbf{Method} & \textbf{$\#$Param} $\downarrow$ & \textbf{Comp.}[$cm$] $\downarrow$ & \textbf{Acc.}[$cm$] $\downarrow$ &  \textbf{Comp.Ratio}[$\%$]$\uparrow$ & \textbf{Acc.Ratio}[$\%$]$\uparrow$  \\
\midrule

\multirow{3}{*}{\texttt{MaiCity}} & \emph{VDBFusion}~\cite{vizzo2022vdbfusion} & $\backslash$ &27.33&\textbf{1.36} &78.12&\textbf{99.13}\\
& \emph{SHINE-Mapping}~\cite{zhong2023shine} &$4.53\times10^{6}$&3.34&1.66&95.43&97.09\\
& Ours &$\mathbf{1.27 \times 10^{6}}$&\textbf{2.68}&1.52&\textbf{97.27}&97.60\\

\midrule

\multirow{3}{*}{\texttt{NewerCollege}} & \emph{VDBFusion} & $\backslash$ & 13.20 &\textbf{5.50}& 91.51 &\textbf{98.10}\\
& \emph{SHINE-Mapping}& $1.14 \times 10^{7}$ &\textbf{9.55}& 7.60 &\textbf{94.58}& 91.37\\
& Ours&$\mathbf{1.60 \times 10^{6}}$&9.68&6.72&94.10&93.69\\

\bottomrule

\end{tabular}
\label{tab.dense_mapping}
\end{table*}%

\begin{figure}
    \centering
    \begin{subfigure}[b]{0.23\textwidth}
        \resizebox{\textwidth}{!}{
            \begin{tikzpicture}[gnuplot]
\path (0.000,0.000) rectangle (6.350,6.350);
\gpcolor{rgb color={0.502,0.502,0.502}}
\gpsetlinetype{gp lt axes}
\gpsetdashtype{gp dt axes}
\gpsetlinewidth{1.00}
\draw[gp path] (1.500,1.165)--(5.797,1.165);
\gpcolor{color=gp lt color border}
\gpsetlinetype{gp lt border}
\gpsetdashtype{gp dt solid}
\draw[gp path] (1.500,1.165)--(1.320,1.165);
\node[gp node right] at (1.136,1.165) {$40$};
\gpcolor{rgb color={0.502,0.502,0.502}}
\gpsetlinetype{gp lt axes}
\gpsetdashtype{gp dt axes}
\draw[gp path] (1.500,1.978)--(1.684,1.978);
\draw[gp path] (5.350,1.978)--(5.797,1.978);
\gpcolor{color=gp lt color border}
\gpsetlinetype{gp lt border}
\gpsetdashtype{gp dt solid}
\draw[gp path] (1.500,1.978)--(1.320,1.978);
\node[gp node right] at (1.136,1.978) {$50$};
\gpcolor{rgb color={0.502,0.502,0.502}}
\gpsetlinetype{gp lt axes}
\gpsetdashtype{gp dt axes}
\draw[gp path] (1.500,2.790)--(5.797,2.790);
\gpcolor{color=gp lt color border}
\gpsetlinetype{gp lt border}
\gpsetdashtype{gp dt solid}
\draw[gp path] (1.500,2.790)--(1.320,2.790);
\node[gp node right] at (1.136,2.790) {$60$};
\gpcolor{rgb color={0.502,0.502,0.502}}
\gpsetlinetype{gp lt axes}
\gpsetdashtype{gp dt axes}
\draw[gp path] (1.500,3.603)--(5.797,3.603);
\gpcolor{color=gp lt color border}
\gpsetlinetype{gp lt border}
\gpsetdashtype{gp dt solid}
\draw[gp path] (1.500,3.603)--(1.320,3.603);
\node[gp node right] at (1.136,3.603) {$70$};
\gpcolor{rgb color={0.502,0.502,0.502}}
\gpsetlinetype{gp lt axes}
\gpsetdashtype{gp dt axes}
\draw[gp path] (1.500,4.416)--(5.797,4.416);
\gpcolor{color=gp lt color border}
\gpsetlinetype{gp lt border}
\gpsetdashtype{gp dt solid}
\draw[gp path] (1.500,4.416)--(1.320,4.416);
\node[gp node right] at (1.136,4.416) {$80$};
\gpcolor{rgb color={0.502,0.502,0.502}}
\gpsetlinetype{gp lt axes}
\gpsetdashtype{gp dt axes}
\draw[gp path] (1.500,5.228)--(5.797,5.228);
\gpcolor{color=gp lt color border}
\gpsetlinetype{gp lt border}
\gpsetdashtype{gp dt solid}
\draw[gp path] (1.500,5.228)--(1.320,5.228);
\node[gp node right] at (1.136,5.228) {$90$};
\gpcolor{rgb color={0.502,0.502,0.502}}
\gpsetlinetype{gp lt axes}
\gpsetdashtype{gp dt axes}
\draw[gp path] (1.500,6.041)--(5.797,6.041);
\gpcolor{color=gp lt color border}
\gpsetlinetype{gp lt border}
\gpsetdashtype{gp dt solid}
\draw[gp path] (1.500,6.041)--(1.320,6.041);
\node[gp node right] at (1.136,6.041) {$100$};
\gpcolor{rgb color={0.502,0.502,0.502}}
\gpsetlinetype{gp lt axes}
\gpsetdashtype{gp dt axes}
\draw[gp path] (1.500,1.165)--(1.500,6.041);
\gpcolor{color=gp lt color border}
\gpsetlinetype{gp lt border}
\gpsetdashtype{gp dt solid}
\draw[gp path] (1.500,1.165)--(1.500,0.985);
\node[gp node center,font={\fontsize{7.0pt}{8.4pt}\selectfont}] at (1.500,0.677) {$1$};
\gpcolor{rgb color={0.502,0.502,0.502}}
\gpsetlinetype{gp lt axes}
\gpsetdashtype{gp dt axes}
\draw[gp path] (2.359,1.165)--(2.359,1.345);
\draw[gp path] (2.359,2.176)--(2.359,6.041);
\gpcolor{color=gp lt color border}
\gpsetlinetype{gp lt border}
\gpsetdashtype{gp dt solid}
\draw[gp path] (2.359,1.165)--(2.359,0.985);
\node[gp node center,font={\fontsize{7.0pt}{8.4pt}\selectfont}] at (2.359,0.677) {$2$};
\gpcolor{rgb color={0.502,0.502,0.502}}
\gpsetlinetype{gp lt axes}
\gpsetdashtype{gp dt axes}
\draw[gp path] (3.219,1.165)--(3.219,1.345);
\draw[gp path] (3.219,2.176)--(3.219,6.041);
\gpcolor{color=gp lt color border}
\gpsetlinetype{gp lt border}
\gpsetdashtype{gp dt solid}
\draw[gp path] (3.219,1.165)--(3.219,0.985);
\node[gp node center,font={\fontsize{7.0pt}{8.4pt}\selectfont}] at (3.219,0.677) {$3$};
\gpcolor{rgb color={0.502,0.502,0.502}}
\gpsetlinetype{gp lt axes}
\gpsetdashtype{gp dt axes}
\draw[gp path] (4.078,1.165)--(4.078,1.345);
\draw[gp path] (4.078,2.176)--(4.078,6.041);
\gpcolor{color=gp lt color border}
\gpsetlinetype{gp lt border}
\gpsetdashtype{gp dt solid}
\draw[gp path] (4.078,1.165)--(4.078,0.985);
\node[gp node center,font={\fontsize{7.0pt}{8.4pt}\selectfont}] at (4.078,0.677) {$4$};
\gpcolor{rgb color={0.502,0.502,0.502}}
\gpsetlinetype{gp lt axes}
\gpsetdashtype{gp dt axes}
\draw[gp path] (4.938,1.165)--(4.938,1.345);
\draw[gp path] (4.938,2.176)--(4.938,6.041);
\gpcolor{color=gp lt color border}
\gpsetlinetype{gp lt border}
\gpsetdashtype{gp dt solid}
\draw[gp path] (4.938,1.165)--(4.938,0.985);
\node[gp node center,font={\fontsize{7.0pt}{8.4pt}\selectfont}] at (4.938,0.677) {$5$};
\gpcolor{rgb color={0.502,0.502,0.502}}
\gpsetlinetype{gp lt axes}
\gpsetdashtype{gp dt axes}
\draw[gp path] (5.797,1.165)--(5.797,6.041);
\gpcolor{color=gp lt color border}
\gpsetlinetype{gp lt border}
\gpsetdashtype{gp dt solid}
\draw[gp path] (5.797,1.165)--(5.797,0.985);
\node[gp node center,font={\fontsize{7.0pt}{8.4pt}\selectfont}] at (5.797,0.677) {$6$};
\node[gp node center,rotate=-270] at (0.292,3.603) {\emph{Completion Ratio}[\%]};
\node[gp node center] at (3.648,0.215) {every $n_{s}$ frame};
\gpcolor{rgb color={0.000,0.000,1.000}}
\gpsetdashtype{gp dt 4}
\gpsetlinewidth{2.00}
\draw[gp path] (1.500,4.263)--(2.359,3.887)--(3.219,3.533)--(4.078,3.267)--(4.938,2.679)%
  --(5.797,2.639);
\gpsetpointsize{8.00}
\gp3point{gp mark 12}{}{(1.500,4.263)}
\gp3point{gp mark 12}{}{(2.359,3.887)}
\gp3point{gp mark 12}{}{(3.219,3.533)}
\gp3point{gp mark 12}{}{(4.078,3.267)}
\gp3point{gp mark 12}{}{(4.938,2.679)}
\gp3point{gp mark 12}{}{(5.797,2.639)}
\gpcolor{rgb color={1.000,0.647,0.000}}
\gpsetdashtype{gp dt 1}
\draw[gp path] (1.500,5.670)--(2.359,5.539)--(3.219,5.404)--(4.078,5.298)--(4.938,5.298)%
  --(5.797,5.110);
\gp3point{gp mark 6}{}{(1.500,5.670)}
\gp3point{gp mark 6}{}{(2.359,5.539)}
\gp3point{gp mark 6}{}{(3.219,5.404)}
\gp3point{gp mark 6}{}{(4.078,5.298)}
\gp3point{gp mark 6}{}{(4.938,5.298)}
\gp3point{gp mark 6}{}{(5.797,5.110)}
\gpcolor{rgb color={1.000,0.000,0.000}}
\draw[gp path] (1.500,5.819)--(2.359,5.713)--(3.219,5.404)--(4.078,5.438)--(4.938,5.320)%
  --(5.797,5.293);
\gp3point{gp mark 11}{}{(1.500,5.819)}
\gp3point{gp mark 11}{}{(2.359,5.713)}
\gp3point{gp mark 11}{}{(3.219,5.404)}
\gp3point{gp mark 11}{}{(4.078,5.438)}
\gp3point{gp mark 11}{}{(4.938,5.320)}
\gp3point{gp mark 11}{}{(5.797,5.293)}
\gpcolor{color=gp lt color border}
\node[gp node left,font={\fontsize{9.0pt}{10.8pt}\selectfont}] at (1.684,2.037) {\emph{VDBFusion}\cite{vizzo2022vdbfusion}};
\gpcolor{rgb color={0.000,0.000,1.000}}
\gpsetlinewidth{1.00}
\gpsetpointsize{6.00}
\gp3point{gp mark 12}{}{(4.762,2.037)}
\gpcolor{color=gp lt color border}
\node[gp node left,font={\fontsize{9.0pt}{10.8pt}\selectfont}] at (1.684,1.760) {\emph{SHINE-Mapping}\cite{zhong2023shine}  };
\gpcolor{rgb color={1.000,0.647,0.000}}
\gp3point{gp mark 6}{}{(4.762,1.760)}
\gpcolor{color=gp lt color border}
\node[gp node left,font={\fontsize{9.0pt}{10.8pt}\selectfont}] at (1.684,1.483) {Ours};
\gpcolor{rgb color={1.000,0.000,0.000}}
\gp3point{gp mark 11}{}{(4.762,1.483)}
\gpdefrectangularnode{gp plot 1}{\pgfpoint{1.500cm}{1.165cm}}{\pgfpoint{5.797cm}{6.041cm}}
\end{tikzpicture}
        }
        \subcaption{\texttt{MaiCity}}
    \end{subfigure}
    \begin{subfigure}[b]{0.23\textwidth}
        \resizebox{\textwidth}{!}{
            \begin{tikzpicture}[gnuplot]
\path (0.000,0.000) rectangle (6.350,6.350);
\gpcolor{rgb color={0.502,0.502,0.502}}
\gpsetlinetype{gp lt axes}
\gpsetdashtype{gp dt axes}
\gpsetlinewidth{1.00}
\draw[gp path] (1.500,1.165)--(5.797,1.165);
\gpcolor{color=gp lt color border}
\gpsetlinetype{gp lt border}
\gpsetdashtype{gp dt solid}
\draw[gp path] (1.500,1.165)--(1.320,1.165);
\node[gp node right] at (1.136,1.165) {$40$};
\gpcolor{rgb color={0.502,0.502,0.502}}
\gpsetlinetype{gp lt axes}
\gpsetdashtype{gp dt axes}
\draw[gp path] (1.500,1.978)--(5.797,1.978);
\gpcolor{color=gp lt color border}
\gpsetlinetype{gp lt border}
\gpsetdashtype{gp dt solid}
\draw[gp path] (1.500,1.978)--(1.320,1.978);
\node[gp node right] at (1.136,1.978) {$50$};
\gpcolor{rgb color={0.502,0.502,0.502}}
\gpsetlinetype{gp lt axes}
\gpsetdashtype{gp dt axes}
\draw[gp path] (1.500,2.790)--(5.797,2.790);
\gpcolor{color=gp lt color border}
\gpsetlinetype{gp lt border}
\gpsetdashtype{gp dt solid}
\draw[gp path] (1.500,2.790)--(1.320,2.790);
\node[gp node right] at (1.136,2.790) {$60$};
\gpcolor{rgb color={0.502,0.502,0.502}}
\gpsetlinetype{gp lt axes}
\gpsetdashtype{gp dt axes}
\draw[gp path] (1.500,3.603)--(5.797,3.603);
\gpcolor{color=gp lt color border}
\gpsetlinetype{gp lt border}
\gpsetdashtype{gp dt solid}
\draw[gp path] (1.500,3.603)--(1.320,3.603);
\node[gp node right] at (1.136,3.603) {$70$};
\gpcolor{rgb color={0.502,0.502,0.502}}
\gpsetlinetype{gp lt axes}
\gpsetdashtype{gp dt axes}
\draw[gp path] (1.500,4.416)--(5.797,4.416);
\gpcolor{color=gp lt color border}
\gpsetlinetype{gp lt border}
\gpsetdashtype{gp dt solid}
\draw[gp path] (1.500,4.416)--(1.320,4.416);
\node[gp node right] at (1.136,4.416) {$80$};
\gpcolor{rgb color={0.502,0.502,0.502}}
\gpsetlinetype{gp lt axes}
\gpsetdashtype{gp dt axes}
\draw[gp path] (1.500,5.228)--(5.797,5.228);
\gpcolor{color=gp lt color border}
\gpsetlinetype{gp lt border}
\gpsetdashtype{gp dt solid}
\draw[gp path] (1.500,5.228)--(1.320,5.228);
\node[gp node right] at (1.136,5.228) {$90$};
\gpcolor{rgb color={0.502,0.502,0.502}}
\gpsetlinetype{gp lt axes}
\gpsetdashtype{gp dt axes}
\draw[gp path] (1.500,6.041)--(5.797,6.041);
\gpcolor{color=gp lt color border}
\gpsetlinetype{gp lt border}
\gpsetdashtype{gp dt solid}
\draw[gp path] (1.500,6.041)--(1.320,6.041);
\node[gp node right] at (1.136,6.041) {$100$};
\gpcolor{rgb color={0.502,0.502,0.502}}
\gpsetlinetype{gp lt axes}
\gpsetdashtype{gp dt axes}
\draw[gp path] (1.500,1.165)--(1.500,6.041);
\gpcolor{color=gp lt color border}
\gpsetlinetype{gp lt border}
\gpsetdashtype{gp dt solid}
\draw[gp path] (1.500,1.165)--(1.500,0.985);
\node[gp node center,font={\fontsize{7.0pt}{8.4pt}\selectfont}] at (1.500,0.677) {3};
\gpcolor{rgb color={0.502,0.502,0.502}}
\gpsetlinetype{gp lt axes}
\gpsetdashtype{gp dt axes}
\draw[gp path] (2.359,1.165)--(2.359,1.345)--(2.359,6.041);
\gpcolor{color=gp lt color border}
\gpsetlinetype{gp lt border}
\gpsetdashtype{gp dt solid}
\draw[gp path] (2.359,1.165)--(2.359,0.985);
\node[gp node center,font={\fontsize{7.0pt}{8.4pt}\selectfont}] at (2.359,0.677) {6};
\gpcolor{rgb color={0.502,0.502,0.502}}
\gpsetlinetype{gp lt axes}
\gpsetdashtype{gp dt axes}
\draw[gp path] (3.219,1.165)--(3.219,6.041);
\gpcolor{color=gp lt color border}
\gpsetlinetype{gp lt border}
\gpsetdashtype{gp dt solid}
\draw[gp path] (3.219,1.165)--(3.219,0.985);
\node[gp node center,font={\fontsize{7.0pt}{8.4pt}\selectfont}] at (3.219,0.677) {9};
\gpcolor{rgb color={0.502,0.502,0.502}}
\gpsetlinetype{gp lt axes}
\gpsetdashtype{gp dt axes}
\draw[gp path] (4.078,1.165)--(4.078,6.041);
\gpcolor{color=gp lt color border}
\gpsetlinetype{gp lt border}
\gpsetdashtype{gp dt solid}
\draw[gp path] (4.078,1.165)--(4.078,0.985);
\node[gp node center,font={\fontsize{7.0pt}{8.4pt}\selectfont}] at (4.078,0.677) {12};
\gpcolor{rgb color={0.502,0.502,0.502}}
\gpsetlinetype{gp lt axes}
\gpsetdashtype{gp dt axes}
\draw[gp path] (4.938,1.165)--(4.938,6.041);
\gpcolor{color=gp lt color border}
\gpsetlinetype{gp lt border}
\gpsetdashtype{gp dt solid}
\draw[gp path] (4.938,1.165)--(4.938,0.985);
\node[gp node center,font={\fontsize{7.0pt}{8.4pt}\selectfont}] at (4.938,0.677) {15};
\gpcolor{rgb color={0.502,0.502,0.502}}
\gpsetlinetype{gp lt axes}
\gpsetdashtype{gp dt axes}
\draw[gp path] (5.797,1.165)--(5.797,6.041);
\gpcolor{color=gp lt color border}
\gpsetlinetype{gp lt border}
\gpsetdashtype{gp dt solid}
\draw[gp path] (5.797,1.165)--(5.797,0.985);
\node[gp node center,font={\fontsize{7.0pt}{8.4pt}\selectfont}] at (5.797,0.677) {18};
\node[gp node center,rotate=-270] at (0.292,3.603) {\emph{Completion Ratio}[\%]};
\node[gp node center] at (3.648,0.215) {every $n_{s}$ frame};
\gpcolor{rgb color={0.000,0.000,1.000}}
\gpsetdashtype{gp dt 4}
\gpsetlinewidth{2.00}
\draw[gp path] (1.500,5.351)--(2.359,5.511)--(3.219,4.863)--(4.078,4.570)--(4.938,4.256)%
  --(5.797,3.887);
\gpsetpointsize{8.00}
\gp3point{gp mark 12}{}{(1.500,5.351)}
\gp3point{gp mark 12}{}{(2.359,5.511)}
\gp3point{gp mark 12}{}{(3.219,4.863)}
\gp3point{gp mark 12}{}{(4.078,4.570)}
\gp3point{gp mark 12}{}{(4.938,4.256)}
\gp3point{gp mark 12}{}{(5.797,3.887)}
\gpcolor{rgb color={1.000,0.647,0.000}}
\gpsetdashtype{gp dt 1}
\draw[gp path] (1.500,5.601)--(2.359,5.548)--(3.219,5.535)--(4.078,5.522)--(4.938,5.518)%
  --(5.797,5.516);
\gp3point{gp mark 6}{}{(1.500,5.601)}
\gp3point{gp mark 6}{}{(2.359,5.548)}
\gp3point{gp mark 6}{}{(3.219,5.535)}
\gp3point{gp mark 6}{}{(4.078,5.522)}
\gp3point{gp mark 6}{}{(4.938,5.518)}
\gp3point{gp mark 6}{}{(5.797,5.516)}
\gpcolor{rgb color={1.000,0.000,0.000}}
\draw[gp path] (1.500,5.562)--(2.359,5.558)--(3.219,5.534)--(4.078,5.477)--(4.938,5.463)%
  --(5.797,5.501);
\gp3point{gp mark 11}{}{(1.500,5.562)}
\gp3point{gp mark 11}{}{(2.359,5.558)}
\gp3point{gp mark 11}{}{(3.219,5.534)}
\gp3point{gp mark 11}{}{(4.078,5.477)}
\gp3point{gp mark 11}{}{(4.938,5.463)}
\gp3point{gp mark 11}{}{(5.797,5.501)}
\gpdefrectangularnode{gp plot 1}{\pgfpoint{1.500cm}{1.165cm}}{\pgfpoint{5.797cm}{6.041cm}}
\end{tikzpicture}
        }
        \subcaption{\texttt{NewerCollege}}
    \end{subfigure}
    \caption{Comparison of \emph{Completion Ratio}[\%] versus the input frame numbers $n_{s}$ on two datasets.
    The threshold is 0.1~m for \texttt{MaiCity} and 0.2~m for \texttt{NewerCollege}.
    As the inputs get sparser, the completion ratio of \emph{VDBFusion} drops significantly, while our method maintains a high completion ratio.
    Though with similar performance, our method uses fewer parameters than \emph{SHINE-Mapping}~(see \cref{fig:parametersVsSparse}).
    }
    \label{fig:sparse_inputs_evaluation}
    \vspace{-0.6cm}
\end{figure}
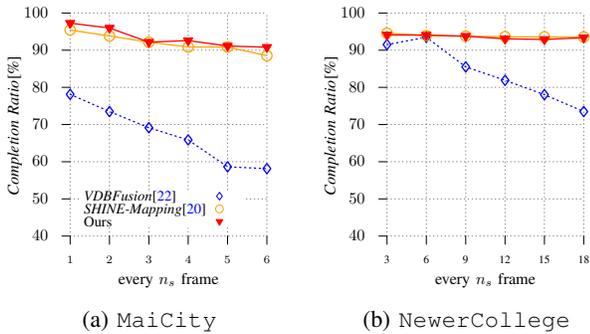

\begin{figure*}
    \centering
    \begin{subfigure}[b]{0.3\textwidth}
        \resizebox{\textwidth}{!}{
            \includegraphics{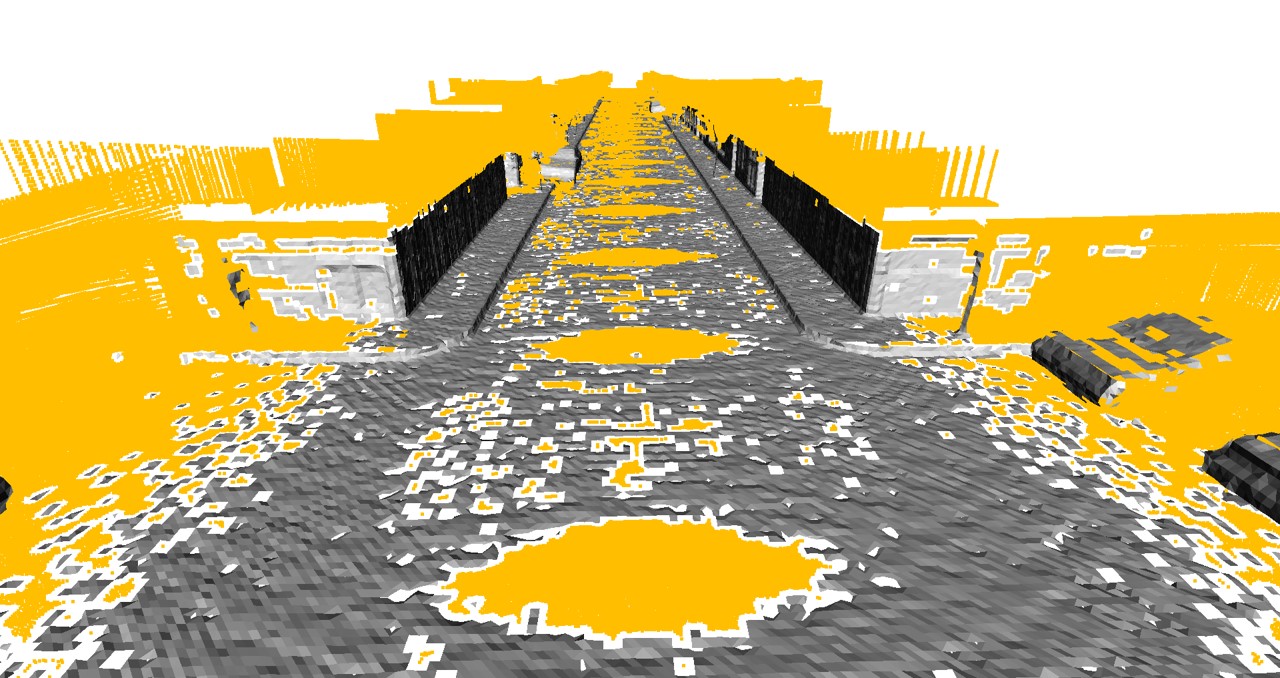}
        }
    \end{subfigure}
    \begin{subfigure}[b]{0.3\textwidth}
        \resizebox{\textwidth}{!}{
            \includegraphics{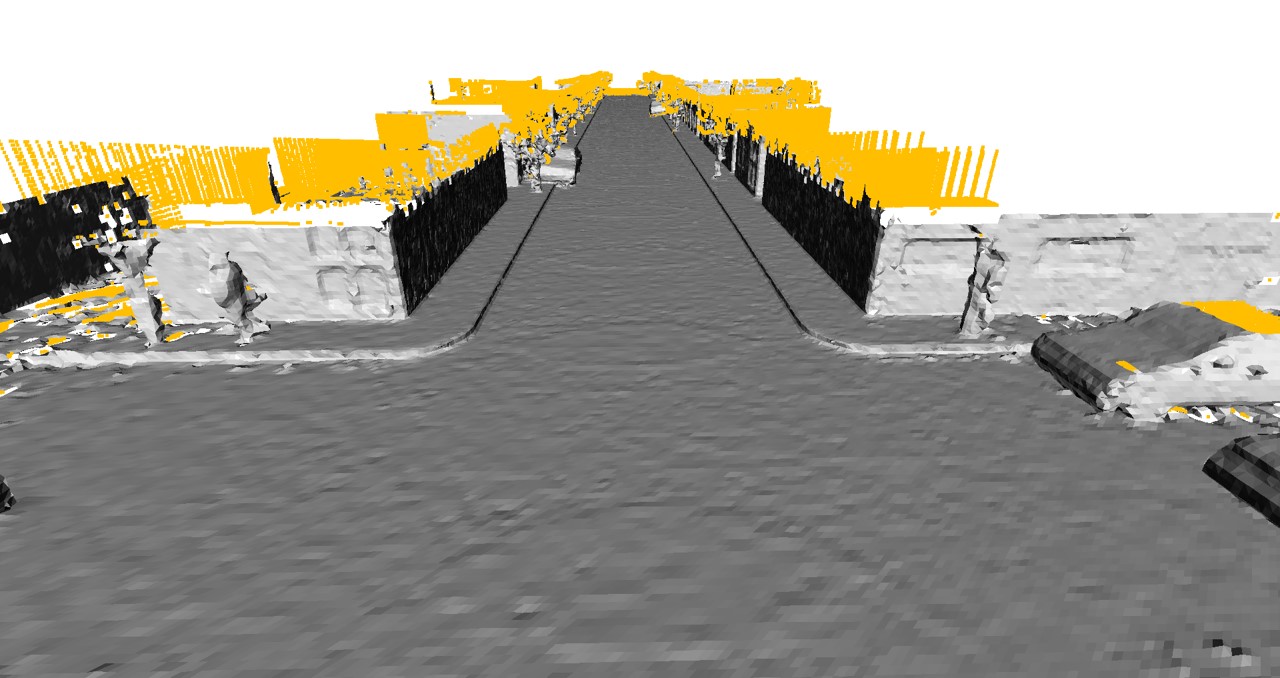}
        }
    \end{subfigure}
    \begin{subfigure}[b]{0.3\textwidth}
        \resizebox{\textwidth}{!}{
            \includegraphics{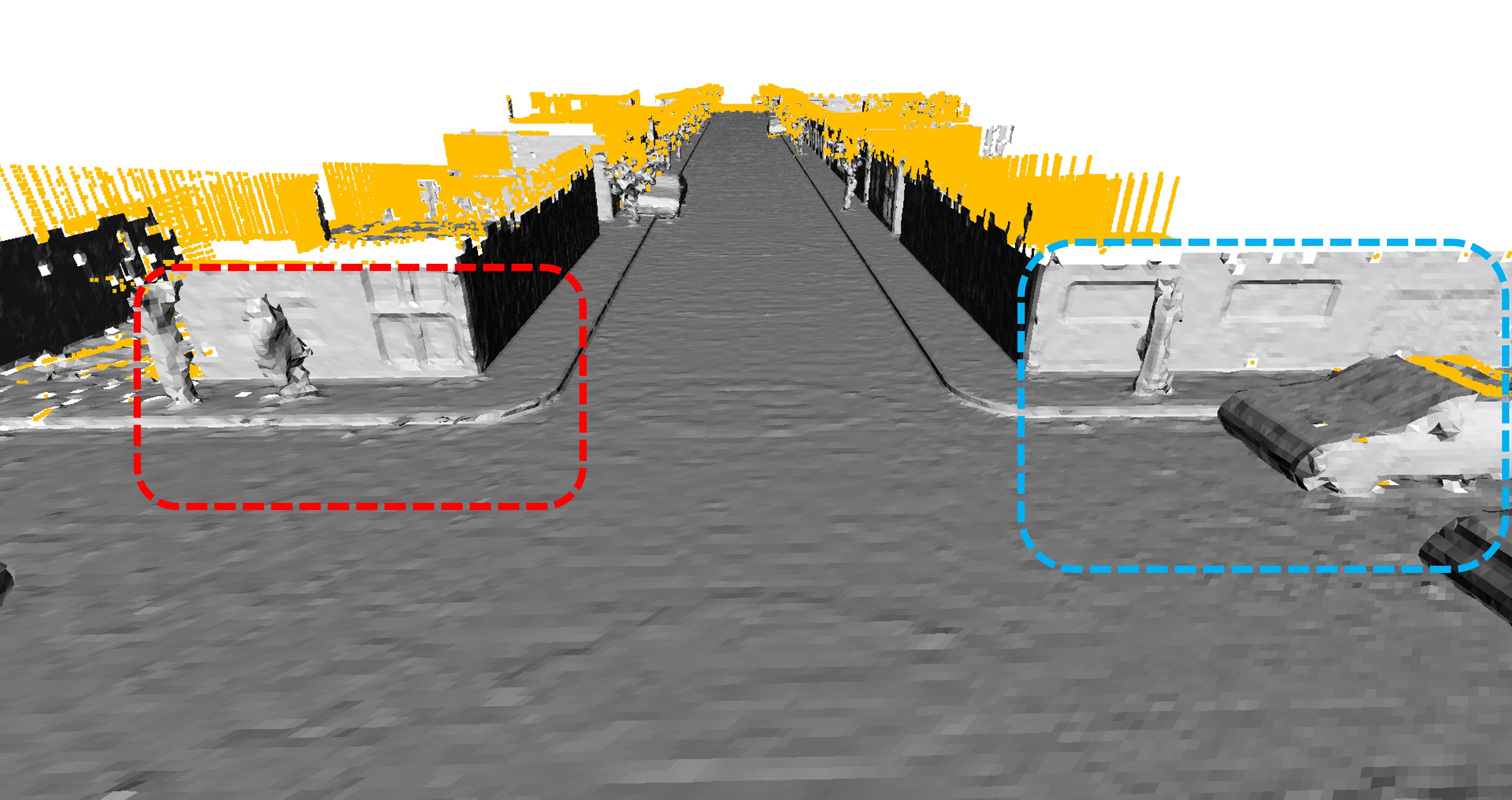}
        }
    \end{subfigure}
    \begin{subfigure}[b]{0.3\textwidth}
        \resizebox{\textwidth}{!}{
            \includegraphics{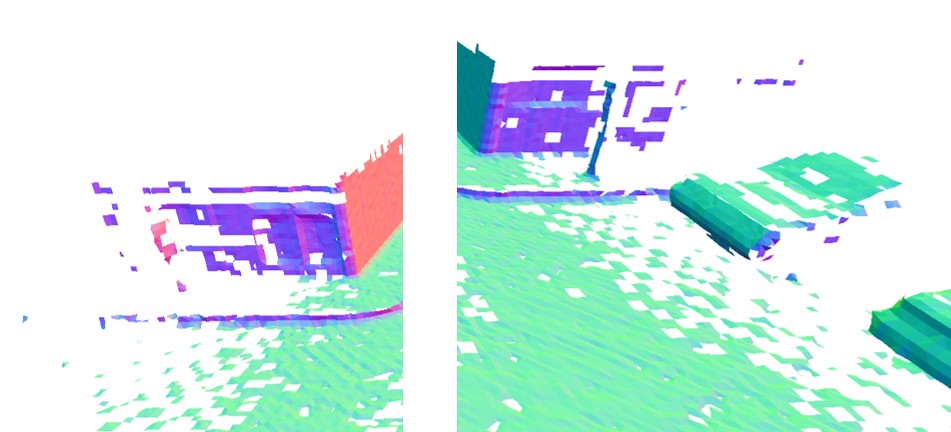}
        }
        \subcaption{\emph{VDBFusion}}
    \end{subfigure}
    \begin{subfigure}[b]{0.3\textwidth}
        \resizebox{\textwidth}{!}{
            \includegraphics{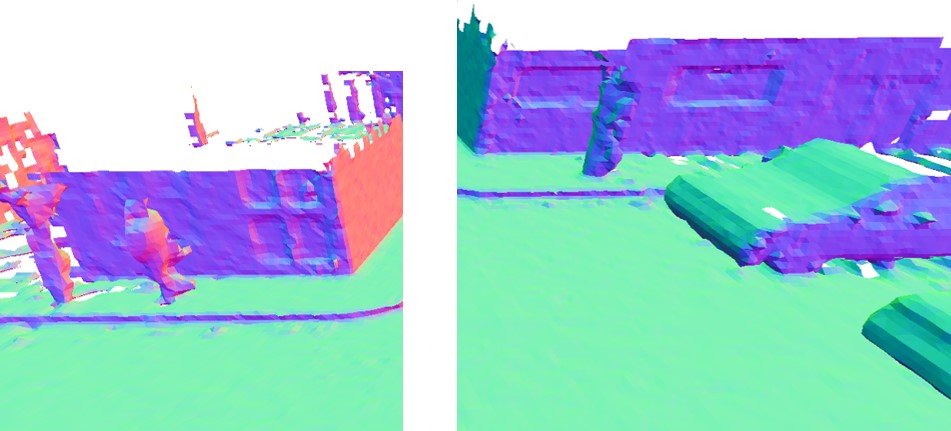}
        }
        \subcaption{\emph{SHINE-Mapping}}
    \end{subfigure}
    \begin{subfigure}[b]{0.3\textwidth}
        \resizebox{\textwidth}{!}{
            \includegraphics{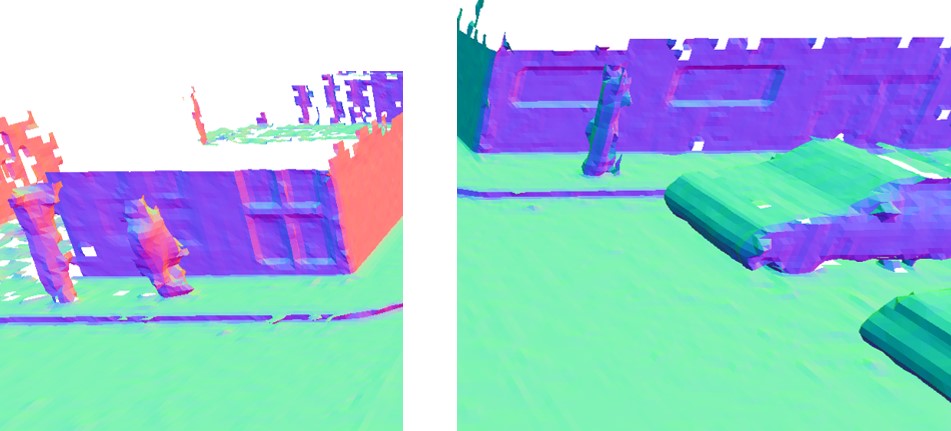}
        }
        \subcaption{Ours}
    \end{subfigure}
    \caption{Qualitative visualization of the map quality on the $\texttt{MaiCity}$ dataset using every 6th frame.
    The first row depicts the difference between the dense ground truth point cloud and the reconstructed mesh; the ground truth points with an error of more than 0.1~m are highlighted in {\color{orange}{orange}}.
    The second row shows zoomed-in images of the dashed areas (indicated in the top-right image).
    When inputs are sparse (e.g., every 6th frame in this case), our method obtains visibly smoother results.}
    \label{fig:completion_ratio_maicity}
    \vspace{-0.3cm}
\end{figure*}

Taking into account that many mapping pipelines use sparse keyframes and not all input frames~\cite{adolfsson-2019-submap,jiang2023h2},
we further evaluate the methods when given sparse input.
We select one frame for every $n_{s}$ frames for reconstruction.
Compared to accuracy, completion can better reveal reconstruction ability.
Focusing on completion-related metrics, one can see in \cref{fig:sparse_inputs_evaluation},
that the neural implicit representation methods maintain a high completion ratio despite sparser inputs.
A visualization of the reconstructed result is shown in \cref{fig:completion_ratio_maicity}, where we highlight the ground truth points with errors larger than 10 cm in {\color{orange}{orange}}.
We can clearly see that there are many ``holes" in the reconstructed mesh of \emph{VDBFusion}, resulting in a low completion ratio.

\begin{figure}
    \centering
    \begin{subfigure}[b]{0.23\textwidth}
        \resizebox{\textwidth}{!}{
            \begin{tikzpicture}[gnuplot]
\path (0.000,0.000) rectangle (6.350,6.350);
\gpcolor{rgb color={0.502,0.502,0.502}}
\gpsetlinetype{gp lt axes}
\gpsetdashtype{gp dt axes}
\gpsetlinewidth{1.00}
\draw[gp path] (1.868,1.165)--(5.797,1.165);
\gpcolor{color=gp lt color border}
\gpsetlinetype{gp lt border}
\gpsetdashtype{gp dt solid}
\draw[gp path] (1.868,1.165)--(1.688,1.165);
\node[gp node right,font={\fontsize{8.0pt}{9.6pt}\selectfont}] at (1.504,1.165) {0e+00};
\gpcolor{rgb color={0.502,0.502,0.502}}
\gpsetlinetype{gp lt axes}
\gpsetdashtype{gp dt axes}
\draw[gp path] (1.868,2.140)--(5.797,2.140);
\gpcolor{color=gp lt color border}
\gpsetlinetype{gp lt border}
\gpsetdashtype{gp dt solid}
\draw[gp path] (1.868,2.140)--(1.688,2.140);
\node[gp node right,font={\fontsize{8.0pt}{9.6pt}\selectfont}] at (1.504,2.140) {1e+06};
\gpcolor{rgb color={0.502,0.502,0.502}}
\gpsetlinetype{gp lt axes}
\gpsetdashtype{gp dt axes}
\draw[gp path] (1.868,3.115)--(5.797,3.115);
\gpcolor{color=gp lt color border}
\gpsetlinetype{gp lt border}
\gpsetdashtype{gp dt solid}
\draw[gp path] (1.868,3.115)--(1.688,3.115);
\node[gp node right,font={\fontsize{8.0pt}{9.6pt}\selectfont}] at (1.504,3.115) {2e+06};
\gpcolor{rgb color={0.502,0.502,0.502}}
\gpsetlinetype{gp lt axes}
\gpsetdashtype{gp dt axes}
\draw[gp path] (1.868,4.091)--(5.797,4.091);
\gpcolor{color=gp lt color border}
\gpsetlinetype{gp lt border}
\gpsetdashtype{gp dt solid}
\draw[gp path] (1.868,4.091)--(1.688,4.091);
\node[gp node right,font={\fontsize{8.0pt}{9.6pt}\selectfont}] at (1.504,4.091) {3e+06};
\gpcolor{rgb color={0.502,0.502,0.502}}
\gpsetlinetype{gp lt axes}
\gpsetdashtype{gp dt axes}
\draw[gp path] (1.868,5.066)--(5.797,5.066);
\gpcolor{color=gp lt color border}
\gpsetlinetype{gp lt border}
\gpsetdashtype{gp dt solid}
\draw[gp path] (1.868,5.066)--(1.688,5.066);
\node[gp node right,font={\fontsize{8.0pt}{9.6pt}\selectfont}] at (1.504,5.066) {4e+06};
\gpcolor{rgb color={0.502,0.502,0.502}}
\gpsetlinetype{gp lt axes}
\gpsetdashtype{gp dt axes}
\draw[gp path] (1.868,6.041)--(5.797,6.041);
\gpcolor{color=gp lt color border}
\gpsetlinetype{gp lt border}
\gpsetdashtype{gp dt solid}
\draw[gp path] (1.868,6.041)--(1.688,6.041);
\node[gp node right,font={\fontsize{8.0pt}{9.6pt}\selectfont}] at (1.504,6.041) {5e+06};
\gpcolor{rgb color={0.502,0.502,0.502}}
\gpsetlinetype{gp lt axes}
\gpsetdashtype{gp dt axes}
\draw[gp path] (1.868,1.165)--(1.868,6.041);
\gpcolor{color=gp lt color border}
\gpsetlinetype{gp lt border}
\gpsetdashtype{gp dt solid}
\draw[gp path] (1.868,1.165)--(1.868,0.985);
\node[gp node center,font={\fontsize{8.0pt}{9.6pt}\selectfont}] at (1.868,0.677) {$1$};
\gpcolor{rgb color={0.502,0.502,0.502}}
\gpsetlinetype{gp lt axes}
\gpsetdashtype{gp dt axes}
\draw[gp path] (2.654,1.165)--(2.654,1.345);
\draw[gp path] (2.654,1.899)--(2.654,6.041);
\gpcolor{color=gp lt color border}
\gpsetlinetype{gp lt border}
\gpsetdashtype{gp dt solid}
\draw[gp path] (2.654,1.165)--(2.654,0.985);
\node[gp node center,font={\fontsize{8.0pt}{9.6pt}\selectfont}] at (2.654,0.677) {$2$};
\gpcolor{rgb color={0.502,0.502,0.502}}
\gpsetlinetype{gp lt axes}
\gpsetdashtype{gp dt axes}
\draw[gp path] (3.440,1.165)--(3.440,1.345);
\draw[gp path] (3.440,1.899)--(3.440,6.041);
\gpcolor{color=gp lt color border}
\gpsetlinetype{gp lt border}
\gpsetdashtype{gp dt solid}
\draw[gp path] (3.440,1.165)--(3.440,0.985);
\node[gp node center,font={\fontsize{8.0pt}{9.6pt}\selectfont}] at (3.440,0.677) {$3$};
\gpcolor{rgb color={0.502,0.502,0.502}}
\gpsetlinetype{gp lt axes}
\gpsetdashtype{gp dt axes}
\draw[gp path] (4.225,1.165)--(4.225,1.345);
\draw[gp path] (4.225,1.899)--(4.225,6.041);
\gpcolor{color=gp lt color border}
\gpsetlinetype{gp lt border}
\gpsetdashtype{gp dt solid}
\draw[gp path] (4.225,1.165)--(4.225,0.985);
\node[gp node center,font={\fontsize{8.0pt}{9.6pt}\selectfont}] at (4.225,0.677) {$4$};
\gpcolor{rgb color={0.502,0.502,0.502}}
\gpsetlinetype{gp lt axes}
\gpsetdashtype{gp dt axes}
\draw[gp path] (5.011,1.165)--(5.011,1.345);
\draw[gp path] (5.011,1.899)--(5.011,6.041);
\gpcolor{color=gp lt color border}
\gpsetlinetype{gp lt border}
\gpsetdashtype{gp dt solid}
\draw[gp path] (5.011,1.165)--(5.011,0.985);
\node[gp node center,font={\fontsize{8.0pt}{9.6pt}\selectfont}] at (5.011,0.677) {$5$};
\gpcolor{rgb color={0.502,0.502,0.502}}
\gpsetlinetype{gp lt axes}
\gpsetdashtype{gp dt axes}
\draw[gp path] (5.797,1.165)--(5.797,6.041);
\gpcolor{color=gp lt color border}
\gpsetlinetype{gp lt border}
\gpsetdashtype{gp dt solid}
\draw[gp path] (5.797,1.165)--(5.797,0.985);
\node[gp node center,font={\fontsize{8.0pt}{9.6pt}\selectfont}] at (5.797,0.677) {$6$};
\node[gp node center,rotate=-270] at (0.292,3.603) {\#Optimizable Parameters};
\node[gp node center] at (3.832,0.215) {every $n_{s}$ frame};
\gpcolor{rgb color={1.000,0.000,0.000}}
\gpsetdashtype{gp dt 1}
\gpsetlinewidth{2.00}
\draw[gp path] (1.868,2.404)--(2.654,2.345)--(3.440,2.306)--(4.225,2.286)--(5.011,2.228)%
  --(5.797,2.228);
\gpsetpointsize{8.00}
\gp3point{gp mark 11}{}{(1.868,2.404)}
\gp3point{gp mark 11}{}{(2.654,2.345)}
\gp3point{gp mark 11}{}{(3.440,2.306)}
\gp3point{gp mark 11}{}{(4.225,2.286)}
\gp3point{gp mark 11}{}{(5.011,2.228)}
\gp3point{gp mark 11}{}{(5.797,2.228)}
\gpcolor{rgb color={1.000,0.647,0.000}}
\draw[gp path] (1.868,5.583)--(2.654,5.261)--(3.440,5.046)--(4.225,4.842)--(5.011,4.715)%
  --(5.797,4.637);
\gp3point{gp mark 6}{}{(1.868,5.583)}
\gp3point{gp mark 6}{}{(2.654,5.261)}
\gp3point{gp mark 6}{}{(3.440,5.046)}
\gp3point{gp mark 6}{}{(4.225,4.842)}
\gp3point{gp mark 6}{}{(5.011,4.715)}
\gp3point{gp mark 6}{}{(5.797,4.637)}
\gpcolor{color=gp lt color border}
\node[gp node left,font={\fontsize{9.0pt}{10.8pt}\selectfont}] at (2.052,1.760) {Ours};
\gpcolor{rgb color={1.000,0.000,0.000}}
\gpsetlinewidth{1.00}
\gpsetpointsize{6.00}
\gp3point{gp mark 11}{}{(5.130,1.760)}
\gpcolor{color=gp lt color border}
\node[gp node left,font={\fontsize{9.0pt}{10.8pt}\selectfont}] at (2.052,1.483) {\emph{SHINE-Mapping}\cite{zhong2023shine}};
\gpcolor{rgb color={1.000,0.647,0.000}}
\gp3point{gp mark 6}{}{(5.130,1.483)}
\gpdefrectangularnode{gp plot 1}{\pgfpoint{1.868cm}{1.165cm}}{\pgfpoint{5.797cm}{6.041cm}}
\end{tikzpicture}
        }
        \subcaption{\texttt{MaiCity}}
    \end{subfigure}
    \begin{subfigure}[b]{0.23\textwidth}
        \resizebox{\textwidth}{!}{
            \begin{tikzpicture}[gnuplot]
\path (0.000,0.000) rectangle (6.350,6.350);
\gpcolor{rgb color={0.502,0.502,0.502}}
\gpsetlinetype{gp lt axes}
\gpsetdashtype{gp dt axes}
\gpsetlinewidth{1.00}
\draw[gp path] (1.928,1.571)--(5.797,1.571);
\gpcolor{color=gp lt color border}
\gpsetlinetype{gp lt border}
\gpsetdashtype{gp dt solid}
\draw[gp path] (1.928,1.571)--(1.748,1.571);
\node[gp node right,font={\fontsize{8.0pt}{9.6pt}\selectfont}] at (1.564,1.571) {1.0e+06};
\gpcolor{rgb color={0.502,0.502,0.502}}
\gpsetlinetype{gp lt axes}
\gpsetdashtype{gp dt axes}
\draw[gp path] (1.928,1.165)--(5.797,1.165);
\gpcolor{color=gp lt color border}
\gpsetlinetype{gp lt border}
\gpsetdashtype{gp dt solid}
\draw[gp path] (1.928,1.165)--(1.748,1.165);
\node[gp node right,font={\fontsize{8.0pt}{9.6pt}\selectfont}] at (1.564,1.165) {0.0e+00};
\gpcolor{rgb color={0.502,0.502,0.502}}
\gpsetlinetype{gp lt axes}
\gpsetdashtype{gp dt axes}
\draw[gp path] (1.928,1.978)--(5.797,1.978);
\gpcolor{color=gp lt color border}
\gpsetlinetype{gp lt border}
\gpsetdashtype{gp dt solid}
\draw[gp path] (1.928,1.978)--(1.748,1.978);
\node[gp node right,font={\fontsize{8.0pt}{9.6pt}\selectfont}] at (1.564,1.978) {2.0e+06};
\gpcolor{rgb color={0.502,0.502,0.502}}
\gpsetlinetype{gp lt axes}
\gpsetdashtype{gp dt axes}
\draw[gp path] (1.928,2.790)--(5.797,2.790);
\gpcolor{color=gp lt color border}
\gpsetlinetype{gp lt border}
\gpsetdashtype{gp dt solid}
\draw[gp path] (1.928,2.790)--(1.748,2.790);
\node[gp node right,font={\fontsize{8.0pt}{9.6pt}\selectfont}] at (1.564,2.790) {4.0e+06};
\gpcolor{rgb color={0.502,0.502,0.502}}
\gpsetlinetype{gp lt axes}
\gpsetdashtype{gp dt axes}
\draw[gp path] (1.928,3.603)--(5.797,3.603);
\gpcolor{color=gp lt color border}
\gpsetlinetype{gp lt border}
\gpsetdashtype{gp dt solid}
\draw[gp path] (1.928,3.603)--(1.748,3.603);
\node[gp node right,font={\fontsize{8.0pt}{9.6pt}\selectfont}] at (1.564,3.603) {6.0e+06};
\gpcolor{rgb color={0.502,0.502,0.502}}
\gpsetlinetype{gp lt axes}
\gpsetdashtype{gp dt axes}
\draw[gp path] (1.928,4.416)--(5.797,4.416);
\gpcolor{color=gp lt color border}
\gpsetlinetype{gp lt border}
\gpsetdashtype{gp dt solid}
\draw[gp path] (1.928,4.416)--(1.748,4.416);
\node[gp node right,font={\fontsize{8.0pt}{9.6pt}\selectfont}] at (1.564,4.416) {8.0e+06};
\gpcolor{rgb color={0.502,0.502,0.502}}
\gpsetlinetype{gp lt axes}
\gpsetdashtype{gp dt axes}
\draw[gp path] (1.928,5.228)--(5.797,5.228);
\gpcolor{color=gp lt color border}
\gpsetlinetype{gp lt border}
\gpsetdashtype{gp dt solid}
\draw[gp path] (1.928,5.228)--(1.748,5.228);
\node[gp node right,font={\fontsize{8.0pt}{9.6pt}\selectfont}] at (1.564,5.228) {1.0e+07};
\gpcolor{rgb color={0.502,0.502,0.502}}
\gpsetlinetype{gp lt axes}
\gpsetdashtype{gp dt axes}
\draw[gp path] (1.928,6.041)--(5.797,6.041);
\gpcolor{color=gp lt color border}
\gpsetlinetype{gp lt border}
\gpsetdashtype{gp dt solid}
\draw[gp path] (1.928,6.041)--(1.748,6.041);
\node[gp node right,font={\fontsize{8.0pt}{9.6pt}\selectfont}] at (1.564,6.041) {1.2e+07};
\gpcolor{rgb color={0.502,0.502,0.502}}
\gpsetlinetype{gp lt axes}
\gpsetdashtype{gp dt axes}
\draw[gp path] (1.928,1.165)--(1.928,6.041);
\gpcolor{color=gp lt color border}
\gpsetlinetype{gp lt border}
\gpsetdashtype{gp dt solid}
\draw[gp path] (1.928,1.165)--(1.928,0.985);
\node[gp node center,font={\fontsize{9.0pt}{10.8pt}\selectfont}] at (1.928,0.677) {3};
\gpcolor{rgb color={0.502,0.502,0.502}}
\gpsetlinetype{gp lt axes}
\gpsetdashtype{gp dt axes}
\draw[gp path] (2.702,1.165)--(2.702,6.041);
\gpcolor{color=gp lt color border}
\gpsetlinetype{gp lt border}
\gpsetdashtype{gp dt solid}
\draw[gp path] (2.702,1.165)--(2.702,0.985);
\node[gp node center,font={\fontsize{9.0pt}{10.8pt}\selectfont}] at (2.702,0.677) {6};
\gpcolor{rgb color={0.502,0.502,0.502}}
\gpsetlinetype{gp lt axes}
\gpsetdashtype{gp dt axes}
\draw[gp path] (3.476,1.165)--(3.476,6.041);
\gpcolor{color=gp lt color border}
\gpsetlinetype{gp lt border}
\gpsetdashtype{gp dt solid}
\draw[gp path] (3.476,1.165)--(3.476,0.985);
\node[gp node center,font={\fontsize{9.0pt}{10.8pt}\selectfont}] at (3.476,0.677) {9};
\gpcolor{rgb color={0.502,0.502,0.502}}
\gpsetlinetype{gp lt axes}
\gpsetdashtype{gp dt axes}
\draw[gp path] (4.249,1.165)--(4.249,6.041);
\gpcolor{color=gp lt color border}
\gpsetlinetype{gp lt border}
\gpsetdashtype{gp dt solid}
\draw[gp path] (4.249,1.165)--(4.249,0.985);
\node[gp node center,font={\fontsize{9.0pt}{10.8pt}\selectfont}] at (4.249,0.677) {12};
\gpcolor{rgb color={0.502,0.502,0.502}}
\gpsetlinetype{gp lt axes}
\gpsetdashtype{gp dt axes}
\draw[gp path] (5.023,1.165)--(5.023,5.861)--(5.023,6.041);
\gpcolor{color=gp lt color border}
\gpsetlinetype{gp lt border}
\gpsetdashtype{gp dt solid}
\draw[gp path] (5.023,1.165)--(5.023,0.985);
\node[gp node center,font={\fontsize{9.0pt}{10.8pt}\selectfont}] at (5.023,0.677) {15};
\gpcolor{rgb color={0.502,0.502,0.502}}
\gpsetlinetype{gp lt axes}
\gpsetdashtype{gp dt axes}
\draw[gp path] (5.797,1.165)--(5.797,6.041);
\gpcolor{color=gp lt color border}
\gpsetlinetype{gp lt border}
\gpsetdashtype{gp dt solid}
\draw[gp path] (5.797,1.165)--(5.797,0.985);
\node[gp node center,font={\fontsize{9.0pt}{10.8pt}\selectfont}] at (5.797,0.677) {18};
\node[gp node center] at (3.862,0.215) {every $n_{s}$ frame};
\gpcolor{rgb color={1.000,0.000,0.000}}
\gpsetdashtype{gp dt 1}
\gpsetlinewidth{2.00}
\draw[gp path] (1.928,1.815)--(2.702,1.815)--(3.476,1.803)--(4.249,1.795)--(5.023,1.791)%
  --(5.797,1.783);
\gpsetpointsize{8.00}
\gp3point{gp mark 11}{}{(1.928,1.815)}
\gp3point{gp mark 11}{}{(2.702,1.815)}
\gp3point{gp mark 11}{}{(3.476,1.803)}
\gp3point{gp mark 11}{}{(4.249,1.795)}
\gp3point{gp mark 11}{}{(5.023,1.791)}
\gp3point{gp mark 11}{}{(5.797,1.783)}
\gpcolor{rgb color={1.000,0.647,0.000}}
\draw[gp path] (1.928,5.757)--(2.702,5.513)--(3.476,5.310)--(4.249,5.119)--(5.023,5.058)%
  --(5.797,4.968);
\gp3point{gp mark 6}{}{(1.928,5.757)}
\gp3point{gp mark 6}{}{(2.702,5.513)}
\gp3point{gp mark 6}{}{(3.476,5.310)}
\gp3point{gp mark 6}{}{(4.249,5.119)}
\gp3point{gp mark 6}{}{(5.023,5.058)}
\gp3point{gp mark 6}{}{(5.797,4.968)}
\gpdefrectangularnode{gp plot 1}{\pgfpoint{1.928cm}{1.165cm}}{\pgfpoint{5.797cm}{6.041cm}}
\end{tikzpicture}
        }
        \subcaption{\texttt{NewerCollege}}
    \end{subfigure}
    \caption{Number of learnable parameters versus subsampling frequency given as $n_{s}$, the number of frames after which another frame was selected from the two datasets.
    Our method only needs about $25\%$ and $10\%$ parameters of \emph{SHINE-Mapping} on \texttt{MaiCity} and the \texttt{NewerCollege} dataset, respectively.
    }
    \label{fig:parametersVsSparse}
\end{figure}
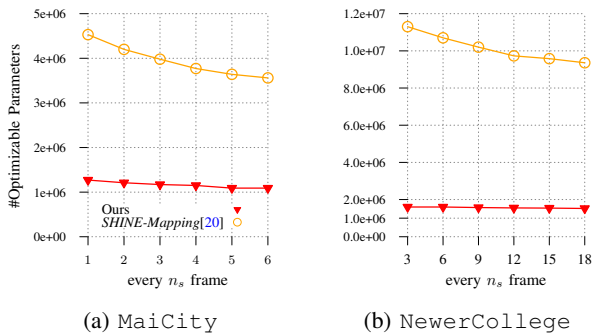

Although \emph{SHINE-Mapping} has a similar completion ratio to our approach,
we show that thanks to our compact and efficient data structure, our method needs substantially fewer parameters and achieves a similar performance.
As demonstrated in \cref{fig:parametersVsSparse}, our method needs about $25\%$ and $10\%$ as many parameters as \emph{SHINE-Mapping} on \texttt{MaiCity} and \texttt{NewerCollege} respectively.
We will discuss more about the memory footprint in the following section (\ref{sec.mem_eff}).
Also, in the second row of \cref{fig:completion_ratio_maicity}, we show that when inputs are sparse, our method can achieve a smoother reconstruction.

\subsection{Memory Footprint} \label{sec.mem_eff}
\begin{figure}
    \centering
    \begin{subfigure}[b]{0.23\textwidth}
        \centering
        \resizebox{\textwidth}{!}{
            \input{figures/map_size_mai_city}
        }
    \end{subfigure}
    \begin{subfigure}[b]{0.24\textwidth}
        \centering
        \resizebox{\textwidth}{!}{
            \includegraphics{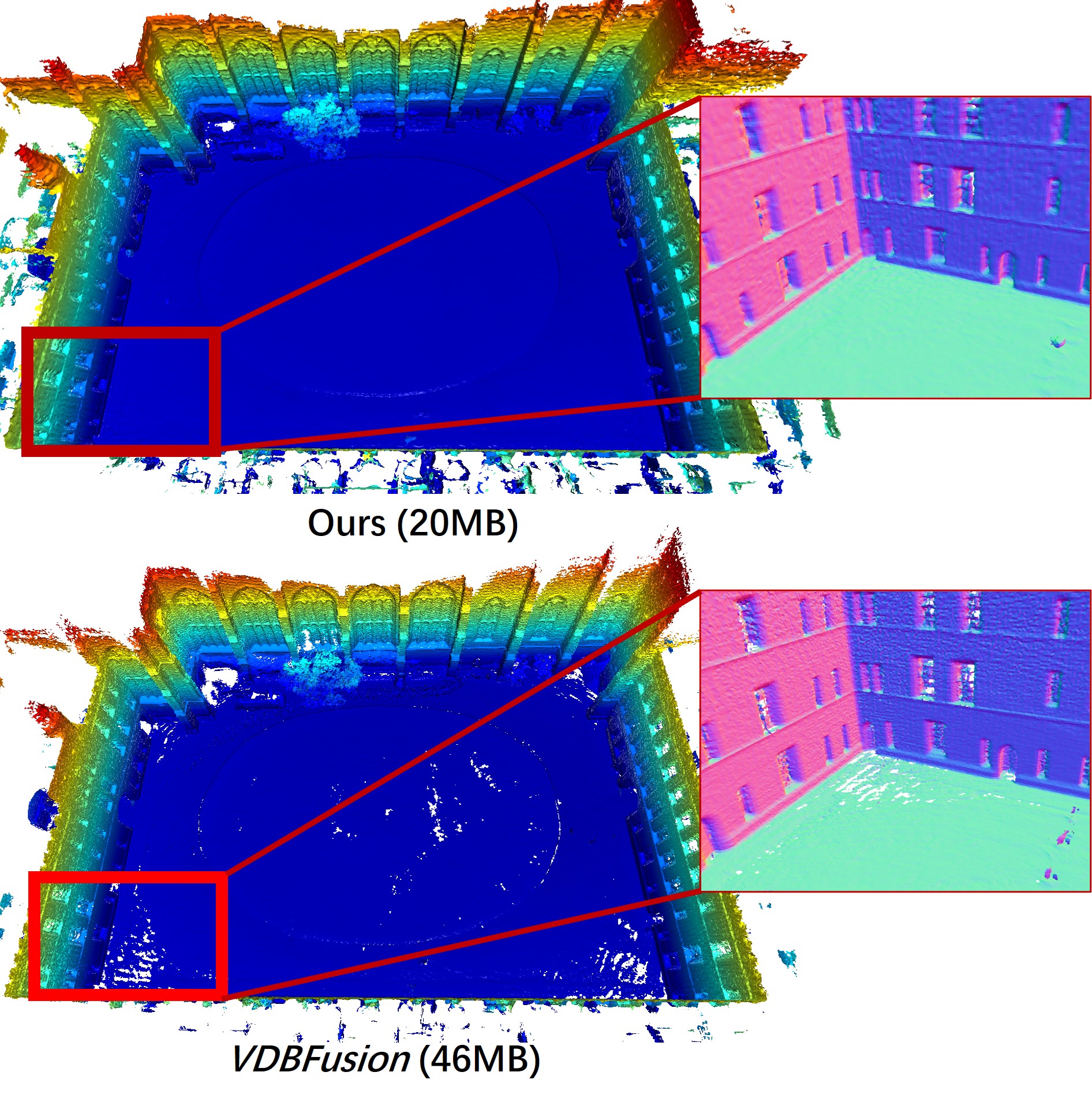}
        }
    \end{subfigure}
    \caption{Memory consumption for different reconstruction methods for the $\texttt{NewerCollege}$ dataset.
    Storing pure point clouds consumes the largest memory usage.
    Our method is much more compact than \emph{SHINE-Mapping}, requiring only $20\%$ of the map size while still achieving comparable results.
    Compared to \emph{VDBFusion}, our method requires less memory but produces a more accurate map; as can be seen in the right column, there are some holes in the map generated by \emph{VDBFusion}.
    }
    \label{fig:memoryFootprint}
\end{figure}

In this section, we analyze the memory usage in the experiment.
For \emph{VDBFusion}, we save its VDB structure as the map representation, which stores TSDF values in grids.
For neural implicit representation methods, we can save the parameters in the model and thus determine the footprint by the size of the saved model, which consists of learned parameters and the MLP decoder parameters.
Fixing the voxel size, \cref{fig:memoryFootprint} shows the memory consumption of the different methods for the \texttt{NewerCollege} dataset.
One can see that tri-quadtrees is the most efficient method with respect to memory and the number of input scans: the point cloud map stores dense 3D points and uses the most memory, and tri-quadtrees requires only 20\% of \emph{SHINE-Mapping} memory usage due to \emph{SHINE-Mapping} storing learned features in 3D voxels.
Tri-quadtrees can use less memory but achieve better reconstruction than \emph{VDBFusion}.
Overall, tri-quadtrees achieves comparable map quality to the state-of-art while being more memory efficient.

\subsection{Ablation Study:~Positional Encoding} \label{sec.abla}
We have performed an ablation study to demonstrate the relative performance of our proposed combined feature encoding vs. the tri-quadtrees features and Fourier feature positional encoding separately.

\begin{table}
\centering
\captionsetup{position=top}
\caption{Ablation study on the different encoding methods.
The evaluation is conducted on \texttt{MaiCity}.
The full model combining learnable features and positional encoding achieves the best performance overall.
Using positional encoding alone can result in an over-smooth result and generate stripe artifacts (see \Cref{fig:ablation_full_vs_pos}).
}
\begin{tabular}{lccc}
\toprule
& w/o Pos.enc. & w/o Fea & Full \\
\midrule
\textbf{Comp.}[cm] $\downarrow$&8.05&9.27&\textbf{7.88}\\
\textbf{Comp.Ratio}[$\%$]$\uparrow$&90.95&89.44&\textbf{91.10}\\
\textbf{Acc.}[cm] $\downarrow$&3.07&2.95&\textbf{2.60}\\
\textbf{Acc.Ratio}[$\%$]$\uparrow$&94.71&\textbf{96.30}&96.25\\
\bottomrule
\label{tab:abliation_study}
\end{tabular}
\vspace{-0.5cm}
\end{table}
\begin{figure}
    \centering
    \begin{subfigure}[b]{0.23\textwidth}
        \centering
        \resizebox{\textwidth}{!}{
            \includegraphics{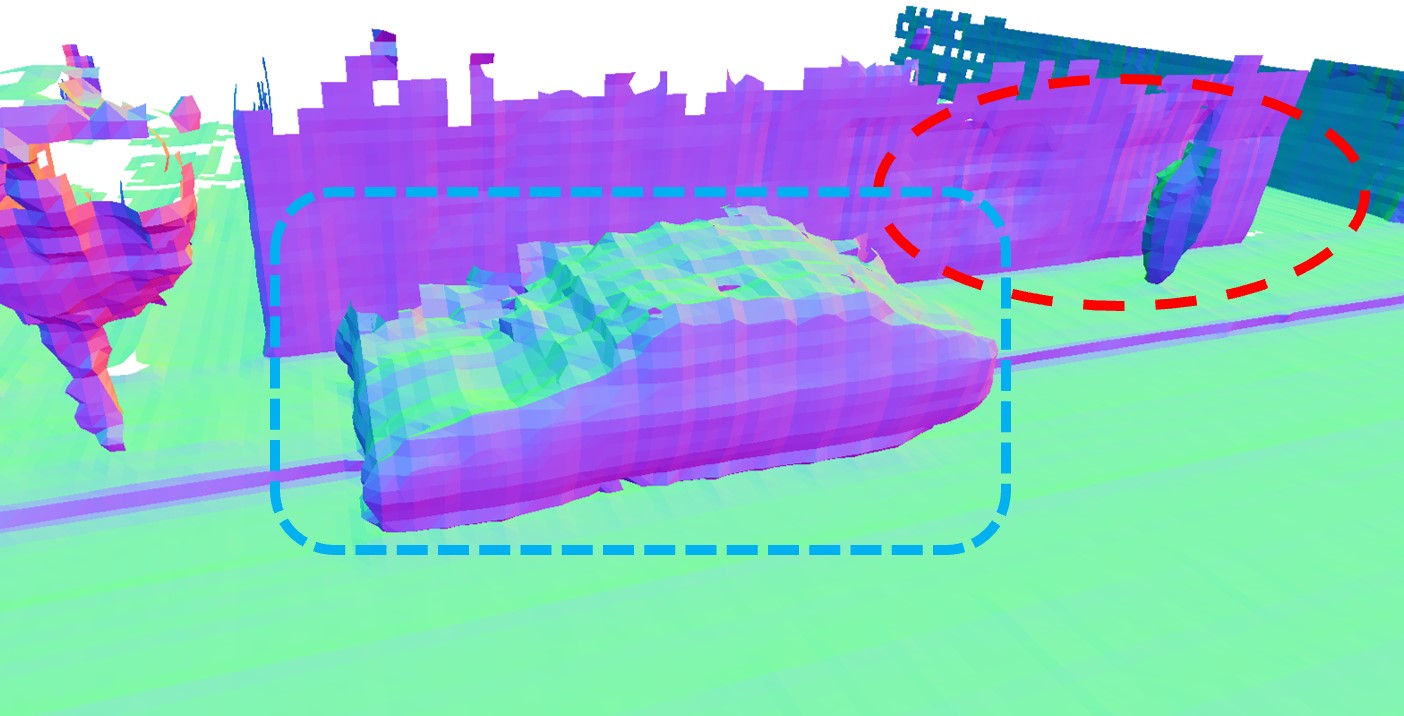}
        }
        \subcaption{Positional encoding only}
    \end{subfigure}
    \begin{subfigure}[b]{0.23\textwidth}
        \centering
        \resizebox{\textwidth}{!}{
            \includegraphics{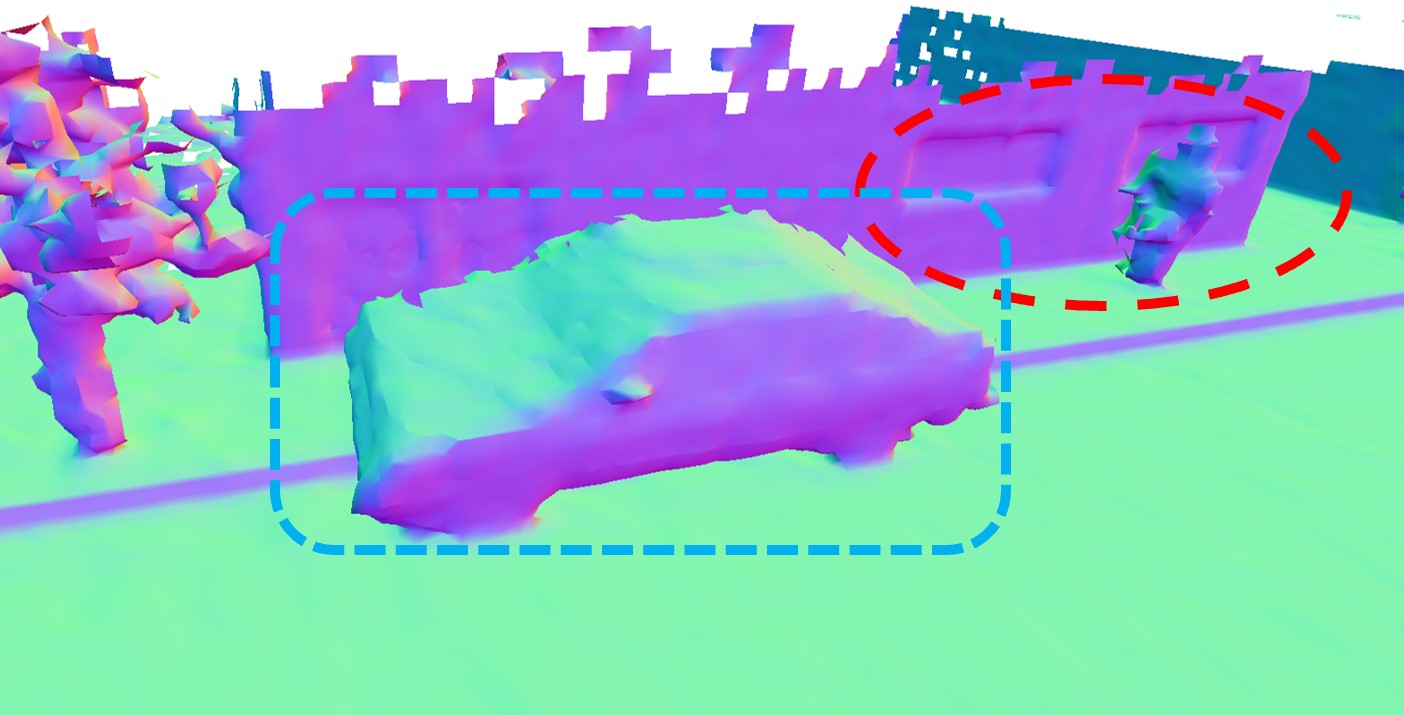}
        }
        \subcaption{Full model}
    \end{subfigure}
    \caption{A comparison of the map quality between using positional-encoding only and the Full model.
    Using positional encoding alone will generate an over-smooth mesh (see the wall in the red circle) and stripe artifacts (see the car in the blue square).
    }
    \label{fig:ablation_full_vs_pos}
\end{figure}

\Cref{tab:abliation_study} illustrates the quantitative evaluation with different encodings on the $\texttt{MaiCity}$ dataset.
Combining both the Fourier feature positional encoding and tri-quadtrees features (namely, ``Full" in \cref{tab:abliation_study}) achieves the best performance overall.
Though using positional encoding alone can achieve a slightly higher accuracy ratio quantitatively, \cref{fig:ablation_full_vs_pos} depicts that the generated mesh lacks details and is with stripe artifacts.

\section{Discussion \& Conclusion}
In this paper, we introduce a novel feature representation method, tri-quadtrees, for neural implicit representation.
Our method combines the advantages of octree sparsity and the compactness of feature plane representation, which requires substantially less memory but achieves better results on completion and competitive results on accuracy.
Thanks to the Fourier feature positional encoding, our method can still achieve a smooth result when the inputs are sparse.

The work in this paper focuses on mapping, which requires given poses.
For future research, we will combine the proposed mapping method with tracking for a neural simultaneous localization and mapping (SLAM) system.
In addition, though a small MLP can be optimized quickly, it lacks enough capacity to fully represent a large-scale environment.
In the future, we will investigate the mapping and fusion with multiple MLPs based on submaps.





\expandafter\def\csname blx@maxbibnames\endcsname{99}%
\printbibliography

\end{document}